\documentclass[letterpaper,journal]{IEEEtran}
\usepackage[letterpaper,margin=1in,includeheadfoot,footskip=12pt]{geometry}
\usepackage[T1]{fontenc}
\usepackage{amsmath,amsfonts}
\usepackage{bm}
\usepackage{array}
\usepackage[caption=false,font=normalsize,labelfont=sf,textfont=sf]{subfig}
\usepackage{textcomp}
\usepackage{stfloats}
\usepackage{url}
\usepackage{verbatim}
\usepackage{graphicx}
\usepackage{cite}
\usepackage{booktabs}
\usepackage{siunitx}
\usepackage{gensymb}
\usepackage{multirow}
\usepackage{xcolor}
\usepackage{pifont}
\graphicspath{{./}}

\newlength{\gtw}
\newcommand{\qhang}{\par\hangindent=\gtw\hangafter=1\relax}
\newcommand{\qcont}{\par\hangindent=\gtw\hangafter=0\relax}
\newcommand{\labline}[2]{\par\begingroup\settowidth{\gtw}{\textbf{#1} }\ifdim\gtw>0.42\linewidth\gtw=0.42\linewidth\fi\hangindent=\gtw\hangafter=1\relax\textbf{#1} #2\par\endgroup}

\usepackage{placeins}

\renewcommand{\arraystretch}{0.9}
\makeatletter
\def\section{\@startsection{section}{1}{\z@}{1.2ex plus 0.5ex minus 0.5ex}%
{0.4ex plus 0.3ex minus 0ex}{\normalfont\normalsize\centering\scshape}}%
\def\subsection{\@startsection{subsection}{2}{\z@}{1.2ex plus 0.5ex minus 0.5ex}%
{0.4ex plus 0.3ex minus 0ex}{\normalfont\normalsize\itshape}}%
\makeatother
\begin{document}

\title{DS-BiCA-Net: Pair-Preserving Cross-Temporal Modeling for Remote Sensing Change Captioning}

\author{Licheng Zhang,~\IEEEmembership{Student Member,~IEEE,}~Naveed Akhtar,~\IEEEmembership{Senior Member,~IEEE,}~and~Siew-Kei Lam,~\IEEEmembership{Senior Member,~IEEE}
\thanks{Licheng Zhang and~Naveed Akhtar are with the School of Computing and Information Systems, the University of Melbourne, Victoria 3010, Australia. (\textit{Corresponding author, Licheng Zhang}, email licheng.zhang@student.unimelb.edu.au)}
\thanks{Siew-Kei Lam is with Nanyang Technological University, Singapore.}}

\markboth{}{}


\makeatletter
\def\@IEEENORMtitlevspace{1.5\baselineskip}
\def\@IEEEMINtitlevspace{1\baselineskip}
\makeatother
\maketitle

\begin{abstract}

Remote sensing image change captioning (RSICC) describes in natural language the semantic changes between two images of the same scene captured at different times. Previous methods commonly merge the pair during repeated processing or combine intra-image and cross-image relations in one operation, weakening source identity and leaving the two temporal directions implicit. Without separate modules for intra-image and cross-temporal processing and a separate path for each direction, the contribution of each part cannot be measured on its own, so it remains unclear which part matters most, and the design of later models has no such evidence to build on. To address these issues, we propose DS-BiCA-Net, which assigns the two kinds of processing to separate modules and keeps the two images as separate streams throughout the encoder. A windowed state-space module models context within each image, and a bidirectional cross-attention module compares the two images along two directed paths that share their projections. The two modules use the same pre-normalisation residual architecture and differ only in their mixers, and the two streams are fused only in the aggregation head, so each module and each direction can be removed and its contribution measured on its own. Controlled experiments on LEVIR-CC and Dubai-CC, reported with paired bootstrap tests under a Holm correction, examine operator interaction, temporal direction and module order, showing that intra-image refinement carries no benefit alone and becomes beneficial only when followed by bidirectional cross-attention, an order that outperforms the reverse at matched parameter count and FLOPs. The bidirectional module also outperforms either temporal direction alone. Our method reaches an $S_m^*$ of $82.65$ on LEVIR-CC, the highest among the published results compared, and $66.72$ under the Dubai-CC protocol without geometric augmentation, the second-highest among the published results compared, rising to $70.14$ and $72.13$ under the two protocols that add geometric augmentation.
\end{abstract}


\section{Introduction}
Changes on the Earth's surface are driven by urbanization, construction and natural disasters. Multitemporal remote sensing imagery has become widely available with the advance of Earth observation technology and provides rich data for monitoring such dynamics over built-up areas~\cite{liu2020building}, forest~\cite{khan2017forest}, and scenes observed by different sensors~\cite{luppino2019unsupervised}.

Remote sensing image change captioning (RSICC) describes the semantic changes between bi-temporal images of one geographic area in natural language~\cite{chouaf2021captioning,hoxha2022change,liu2022remote}, which serves urban and construction monitoring, urbanization analysis, disaster assessment and comprehensive change interpretation~\cite{hoxha2022change,irvinteochat,shi2024multi,liu2024change}. The task is commonly formulated in two stages, a visual understanding stage that compares the multitemporal pair to locate the changing regions and identify their semantic content, followed by a language generation stage that maps the resulting evidence onto a readable sentence.

Recent RSICC research spans attention-based comparison~\cite{liu2022remote,chang2023changes,cai2023interactive,li2024inter,qin2025dgat,xian2025dynamic,sun2024lightweight,zhou2024single,zhang2026lbtcap}, state-space modeling~\cite{liu2024rscama,ma2025cross,cao2026rmnet}, diffusion~\cite{yang2024remote,yu2025diffusion,sun2025mask}, and foundation models~\cite{liu2024change,deng2025changechat,irvinteochat,wang2024ccexpert}. All of them depend on reliable visual comparison, because a powerful decoder can produce fluent but unsupported statements from poorly aligned evidence. Additional masks, detection targets, multitask supervision, or external priors further change the information available during learning~\cite{li2025cd4c,li2024detection,shi2024multi,zhu2024semantic,yang2025enhancing,hang2025text}, which separates such settings from caption-only learning. Within caption-only encoders, two limitations in how the comparison is organised remain.

\begin{figure}[t]
\centering
\includegraphics[width=\columnwidth]{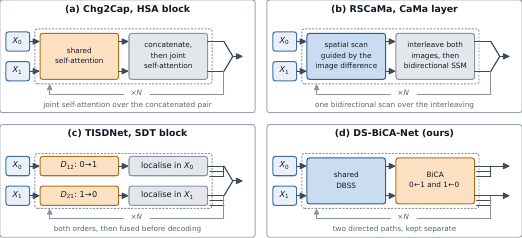}
\caption{Comparison interfaces of Chg2Cap~\cite{chang2023changes}, RSCaMa~\cite{liu2024rscama}, TISDNet~\cite{li2024inter} and DS-BiCA-Net. $X_0$ and $X_1$ are the features of the earlier and the later image of the pair, across which the change is described. Each block is repeated $N$ times.}
\label{fig:interfaces}
\vspace{-2pt}
\end{figure}

The first concerns the temporal pair itself. Many methods merge the two images during repeated processing, through a difference, a concatenation, or a jointly processed sequence~\cite{liu2022remote,chang2023changes,cao2026rmnet}, or combine intra-image and cross-image relations in one operation. Chg2Cap~\cite{chang2023changes}, for example, applies a joint self-attention to the concatenated pair, which makes the two kinds of relation compete within one softmax. TISDNet~\cite{li2024inter} comes closest to a pair-preserving design. It keeps two ordered branches, one localising change in the first image from the difference taken in the forward temporal order and the other in the second image from the difference taken in the reverse order, and fuses the two only before decoding. Each branch, however, is built directly from a feature difference without first refining either image. As a result, the comparison never draws on spatial context within the images it compares. Figure~\ref{fig:interfaces} contrasts these interfaces with ours. Once the pair is merged, the source of an activation is no longer identifiable, the two temporal directions become implicit, and intra-image and cross-temporal contributions cannot be removed separately without changing the interface. Direction matters for the caption, since an object that appears and one that disappears call for different words.

The second concerns the state-space line, which has not adapted its scan to the nature of images. Mamba~\cite{gu2024mamba} was designed for one-dimensional sequences, and its causal scan exposes each token only to the tokens before it in a chosen order. An image has no such order, and flattening its grid gives every location a one-sided context whose extent depends on where the scan happens to start. Existing Mamba-based RSICC methods inherit the causal scan and further ask it to perform temporal comparison. RSCaMa~\cite{liu2024rscama} scans an interleaved bi-temporal sequence with a spatial operator gated by the bi-temporal difference, MTI-CC~\cite{ma2025cross} alternates bidirectional scans over an interleaved sequence with cross-attention to a difference feature, and MTH-Net~\cite{ma2025mth} scans a concatenated temporal sequence and fuses it with such cross-attention through broadcast channel weights, while RMNet~\cite{cao2026rmnet} collapses the pair into a single difference stream. The result then depends on a scan order across time that carries no semantic meaning, and every piece of cross-image evidence must pass through a recurrent state of bounded size, which poorly suits a change confined to a small region.

To address both limitations, we assign each role to the operator that suits it and keep the pair explicit throughout the encoder. Intra-image context is modeled by a dual-branch selective-scan (DBSS) module adopted from~\cite{hatamizadeh2025mambavision}. The module projects the input features and divides them into two branches. One branch applies a selective scan, the other applies only a convolution without scanning, and the outputs of the two branches are concatenated. Every token draws on both its preceding and its following neighbours. The symmetric branch bypasses the scan and compensates for content lost to its sequential constraint~\cite{hatamizadeh2025mambavision}, while the scan branch still supplies content-dependent propagation. The module operates inside fixed windows and preserves feature width and spatial resolution. Cross-temporal comparison is performed by a bidirectional cross-attention (BiCA) module. The same cross-attention, with shared parameters, is applied twice. In the first pass the features of the first image serve as queries and those of the second image as keys and values, and in the second pass the two roles are reversed. Each query can therefore attend to every token of the other image, no information has to pass through a recurrent state of fixed size, no scan order is imposed across time, and the outputs of the two directions remain separate. Two alternatives suit the intra-image role less well. A multi-directional full-grid scan~\cite{cao2026rmnet} would add long-range scan paths, although BiCA already gives every query a global reach, and self-attention within each image~\cite{chang2023changes} would use the same mechanism as the subsequent cross-temporal attention. We name the resulting network DS-BiCA-Net after its dual-branch scan (DS) and BiCA modules. Following prior work~\cite{liu2022remote,chang2023changes,cai2023interactive,li2024inter,sun2024lightweight}, a shared ResNet101~\cite{he2016resnet} extracts both temporal images, after which every DS-BiCA block, a stack of windowed DBSS layers followed by one BiCA layer, accepts and returns two explicit streams that are not collapsed into a single representation until the aggregation head. Spatial and temporal contributions can therefore be removed and measured separately, while block depth and within-block capacity remain free to vary.

To sum up, our main contributions are summarised as follows.
\begin{itemize}
    \setlength{\itemsep}{0pt}
    \item We propose a pair-preserving encoder for RSICC. Every block accepts and returns the two temporal streams, and the streams are fused only in the aggregation head, keeping the source and direction of temporal evidence identifiable throughout the encoder.
    \item We assign intra-image and cross-temporal modeling to different operators. A stack of windowed DBSS modules models the spatial context within each image separately with shared parameters. A BiCA module then compares the two images in both directions using the same weights. Every query thereby reaches all tokens of the other image, and the outputs of the two directions remain separate. Either operator can be removed without changing the interface.
    \item We conduct controlled experiments on LEVIR-CC and Dubai-CC, including a factorial ablation of the two operators, single-direction controls at matched parameter count, and a module-order comparison at matched cost. The ablations show DBSS helping only when BiCA follows it, and the principal contrasts are tested with paired bootstrap under Holm correction.
    \item DS-BiCA-Net reaches 82.65 on the composite metric of LEVIR-CC, the highest among the compared published methods, and 66.72, 70.14 and 72.13 under the three Dubai-CC protocols. Qualitative visualisations illustrate accurate descriptions across diverse changes and scale-specific attention patterns.
\end{itemize}


\section{Related Work}\label{related}
\subsection{Four RSICC Paradigms}
RSICC methods can be broadly categorised into four lines of work, namely attention-based, state-space, diffusion and vision-language methods. Attention-based methods use differences, fused pairs, or joint attention to locate changes~\cite{liu2022remote,chang2023changes,cai2023interactive}. The paradigm has expanded through multi-scale fusion~\cite{xu2024mfrnet,chen2024multi}, selective attention~\cite{qin2025dgat,yang2025enhancing,li2026exploring,gong2026stand,wang2025change,sun2025scene}, and efficiency-oriented designs~\cite{sun2024lightweight,zhou2024single,zhang2026lbtcap,sun2026scnet,wang2025lightweight}. State-space methods replace or combine attention with temporal and spatial scans~\cite{liu2024rscama,ma2025ihm,ma2025cross,wang2025remote,cao2026rmnet,ma2025mth}. Diffusion methods reformulate caption generation or change-region refinement as a denoising process~\cite{yang2024remote,yu2025diffusion,sun2025mask}. Vision-language model (VLM) methods connect bi-temporal visual evidence to large pretrained multimodal decoders~\cite{liu2024change,deng2025changechat,irvinteochat,wang2024ccexpert,yildirim2025llava,xue2026towards,yang2026rscc}. Complementary supervision through masks, detection, multitask targets, textual and semantic priors, foundation models, or spectral indices cuts across these four paradigms~\cite{li2025cd4c,li2024detection,shi2024multi,ma2025mth,li2026exploring,zhu2024semantic,yang2025enhancing,hang2025text}. Our method belongs to the state-space line. It models context within each image with state-space modules, compares the two images with cross-attention, and is trained on captions alone without complementary supervision.

\subsection{Pair-Preserving Comparison Interfaces}

An RSICC encoder must both understand each image and compare the two, and existing methods organise these two steps in different ways.

Chg2Cap~\cite{chang2023changes} retains two temporal outputs before fusion. It applies a shared self-attention to each temporal image and then a joint self-attention to their concatenation. Intra-image and cross-image relations thus enter one softmax, and the method does not represent two separate opposite-image query paths. MTH-Net~\cite{ma2025mth} applies Bi-SSM to the concatenation of the two temporal sequences and fuses its output with that of a cross-attention, in which each image queries a difference feature, through broadcast channel weights. MTI-CC~\cite{ma2025cross} alternates Bi-SSM layers over an interleaved bi-temporal sequence with such cross-attention layers. A paired output alone therefore does not determine how temporal exchange is performed. MTH-Net and MTI-CC both exchange information between the two times by scanning a single sequence merged from the two images, which has two consequences. The two temporal directions cannot be separated, and scan order becomes a hyperparameter with no semantic grounding. More importantly, every piece of cross-image evidence must pass through a recurrent state of bounded size, whereas the change to be described often occupies a small region that such a state is ill-suited to carry. Cross-attention has no fixed-state bottleneck and keeps the two directions explicit, which is why we assign the scan to intra-image context and attention to cross-image comparison.

\subsection{Directionality and Temporal Symmetry}
Existing methods differ in how they compare the two images and in which temporal direction the comparison runs. TISDNet~\cite{li2024inter} localises with two ordered differences, then concatenates the branch outputs. RSCaMa~\cite{liu2024rscama} applies bidirectional state-space modeling to an interleaved sequence. Its spatial operator is itself gated by the bi-temporal difference, which entangles intra-image and cross-temporal processing from the first operator onward. DAE~\cite{xian2025dynamic} refines both streams with difference-guided convolution, then updates only the after stream through an expert in which the after stream queries the before stream by cross-attention. RMNet~\cite{cao2026rmnet} restricts the interaction between the two images to their pixel-level difference, whereas BiCA allows each image to attend directly to the other.

\begin{table}[t]
\caption{How existing change-aware encoders process and compare the two images.}
\label{tab:units}
\centering
\fontsize{8.5}{9.8}\selectfont
\setlength{\tabcolsep}{1.5pt}
\begin{tabular*}{\columnwidth}{@{\extracolsep{\fill}}lcccc@{}}
\toprule
Method & \shortstack{Self-image\\mixing} & \shortstack{Pair\\output} & \shortstack{Directed\\paths} & \shortstack{Multi-layer\\aggregation} \\
\midrule
Chg2Cap~\cite{chang2023changes} & \ding{51} & \ding{51} & \ding{55} & \ding{55} \\
RSCaMa~\cite{liu2024rscama} & \ding{51} & \ding{51} & \ding{55} & \ding{55} \\
TISDNet~\cite{li2024inter} & \ding{55} & \ding{51} & \ding{51} & \ding{51} \\
DAE~\cite{xian2025dynamic} & \ding{51} & \ding{51} & \ding{55} & \ding{51} \\
RMNet~\cite{cao2026rmnet} & \ding{51} & \ding{51} & \ding{55} & \ding{55} \\
MTH-Net~\cite{ma2025mth} & \ding{51} & \ding{51} & \ding{55} & \ding{55} \\
MTI-CC~\cite{ma2025cross} & \ding{51} & \ding{51} & \ding{55} & \ding{55} \\
\midrule
DS-BiCA-Net (\textbf{Ours}) & \ding{51} & \ding{51} & \ding{51} & \ding{51} \\
\bottomrule
\end{tabular*}
\vspace{-7pt}
\end{table}
\subsection{Position of DS-BiCA-Net}
Table~\ref{tab:units} summarises the methods discussed above and ours along four properties. Self-image mixing refines each image with its own spatial context before the comparison. Pair output keeps two feature maps, one per image, so the source of each feature remains known. Directed paths compare the images in both directions as two separate mappings, $0\leftarrow1$ and $1\leftarrow0$. Multi-layer aggregation lets the decoder read the output of every block rather than the last block alone. DS-BiCA-Net is the only method in the table with all four properties. It applies self-image refinement followed by shared opposite-image attention. Each block returns two streams, preserving source identity until aggregation. Unlike ordered differences, combined-sequence scans, or one-sided updates, shared BiCA projections form two query paths that differ only in query and source assignment. DBSS acts within streams and BiCA across them, with separate normalisations and residual paths. Training uses captions alone without masks, detection targets, or external priors.

\section{Methodology}\label{method}
\begin{figure*}[t]
\centering
\includegraphics[width=\textwidth]{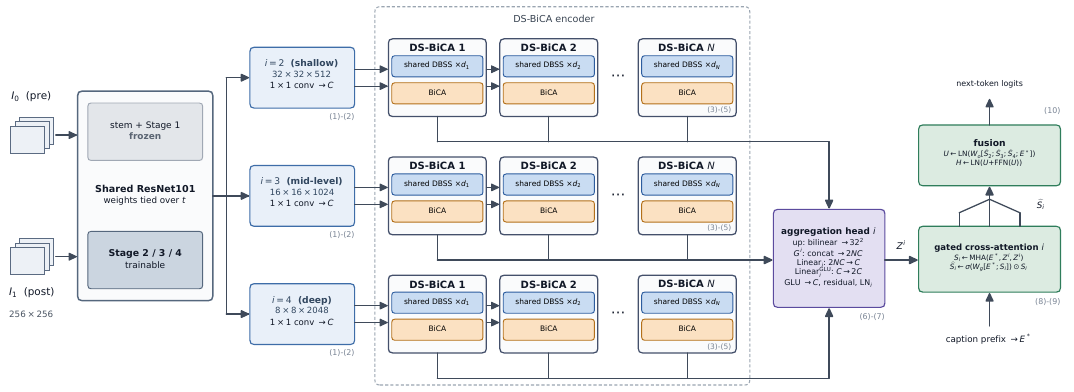}
\caption{Overview of DS-BiCA-Net. A shared ResNet101 extracts features of the earlier image $I_0$ and the later image $I_1$ at three scales. At each scale, a cascade of DS-BiCA blocks first refines each image with its own spatial context (DBSS) and then lets each image attend to the other in both directions (BiCA), keeping the two images as separate streams. The aggregation head fuses the outputs of all blocks into one representation per scale, and the decoder reads the three scales through gated cross-attention to generate the caption word by word.}
\label{fig:mvbicanet}
\vspace{-1pt}
\end{figure*}

Figure~\ref{fig:mvbicanet} presents DS-BiCA-Net. It contains a shared multi-scale backbone, a DS-BiCA encoder, block-wise feature aggregation, and a gated multi-scale caption decoder. The central design matches each operator to the role it suits and keeps the roles apart, rather than asking one operator to serve both. It first builds intra-image context, then exchanges evidence across the temporal pair, and only afterward fuses the two streams for language generation. Delaying fusion keeps the source and direction of an activation identifiable long enough for both query directions and intermediate block outputs to remain observable.

\subsection{Multi-scale Feature Extraction}
Let $I_0$ and $I_1$ denote images of the same scene captured at two times. A ResNet101~\cite{he2016resnet} with shared parameters produces features $R_t^i$ at stages $i\in\{2,3,4\}$ and times $t\in\{0,1\}$. A scale-specific $1\times1$ projection maps the original channel widths $512$, $1024$, and $2048$ to the common width $C$. A learned positional embedding is then added.
\begin{equation}
X_t^i=\phi_i(R_t^i)+P_i, \qquad t\in\{0,1\},\;i\in\{2,3,4\}.
\label{eq:backbone}
\end{equation}
The positional embedding is factorised into learned row and column tables. For a feature map of size $H_i\times W_i$, let $E_i^h\in\mathbb{R}^{H_i\times C/2}$ and $E_i^w\in\mathbb{R}^{W_i\times C/2}$. The embedding at spatial location $(u,v)$ is
\begin{equation}
P_i(u,v)=\operatorname{Concat}\!\left(E_i^w[v],E_i^h[u]\right)\in\mathbb{R}^{C}.
\label{eq:posemb}
\end{equation}
The same $P_i$ of Eq.~\eqref{eq:posemb} is added to both temporal streams.
For $256\times256$ inputs, the three resolutions are $32\times32$, $16\times16$, and $8\times8$.

\begin{figure*}[t]
\centering
\begin{minipage}[c]{0.42297\textwidth}\centering
\includegraphics[width=\linewidth]{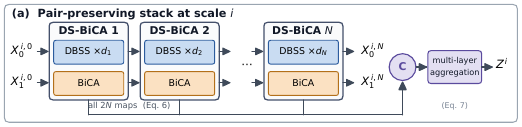}
\includegraphics[width=\linewidth]{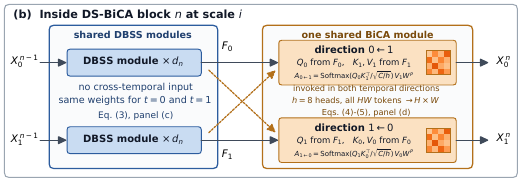}
\end{minipage}\hspace{6pt}%
\begin{minipage}[c]{0.23786\textwidth}\centering
\includegraphics[width=\linewidth]{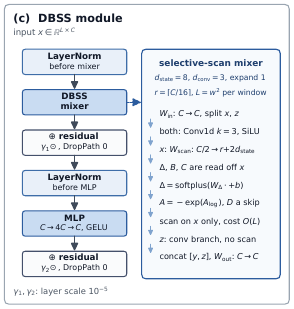}
\end{minipage}\hspace{6pt}%
\begin{minipage}[c]{0.23786\textwidth}\centering
\includegraphics[width=\linewidth]{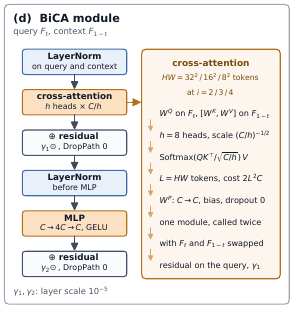}
\end{minipage}
\caption{Pair-preserving DS-BiCA encoder. Here $\oplus$ denotes residual addition and $\odot$ an elementwise product. In panels (c) and (d), $H$, $W$, and $C$ are the feature height, width, and channel count, $L=HW$ is the token count, and $x\in\mathbb{R}^{L\times C}$ is the input sequence. The panels share one module structure and differ only in the mixer, and each footer names the equation that defines a box. Panel (c) is the DBSS module shared by $t=0$ and $t=1$, a pre-norm residual pair around a selective-scan mixer and an MLP. Panel (d) is the BiCA module, a pre-norm cross-attention whose query $F_t$ comes from one stream and keys and values $F_{1-t}$ come from the other, followed by an MLP residual. The layer scales $\gamma_1$ and $\gamma_2$ are those of Eq.~\eqref{eq:bica2}.}
\label{fig:mvbica}
\vspace{-1pt}
\end{figure*}

\subsection{DS-BiCA Encoder}
The projected pair $X_0^i$ and $X_1^i$ of Eq.~\eqref{eq:backbone} then passes through a cascade of $N$ DS-BiCA blocks. Each block operates on a pair and returns a pair. Temporal identity thus survives every block and is given up only at the aggregation. Let $X_t^{i,n}$ denote the feature of stream $t$ at scale $i$ after block $n$, with $X_t^{i,0}=X_t^i$.

Each DS-BiCA block takes the feature pair of the two temporal images and returns an updated pair of the same size. Inside the block, a stack of $d_n$ DBSS modules first models the spatial context of each stream on its own, with the same weights for both streams, and a BiCA module then exchanges evidence between the two streams in both temporal directions. The two kinds of module share one structure with two residual branches, a mixer branch followed by an MLP branch, and differ only in the mixer. For a mixer $g$ applied to a target $U$ with an optional source $S$, the shared structure is defined as
\begin{equation}
\begin{split}
\widehat U&=U+\gamma_{1}^{g}\odot
\mathcal{G}_{g}\!\left(\operatorname{LN}_{q}^{g}(U),
\operatorname{LN}_{s}^{g}(S)\right),\\
\mathcal{T}_{g}(U,S)&=\widehat U+\gamma_{2}^{g}\odot
\operatorname{MLP}_{g}\!\left(\operatorname{LN}_{m}^{g}(\widehat U)\right).
\end{split}
\label{eq:template}
\end{equation}
Here $\gamma_{1}^{g},\gamma_{2}^{g}\in\mathbb{R}^{C}$ are learned layer scales, $\operatorname{LN}_{q}^{g}$ precedes the mixer, and $\operatorname{LN}_{m}^{g}$ precedes the MLP. The source normalisation $\operatorname{LN}_{s}^{g}$ is used only by the BiCA mixer, which takes the source $S$ in addition to the target $U$. Each module owns its normalisations, layer scales, and MLP parameters.

DBSS instantiates Eq.~\eqref{eq:template} with the single-input dual-branch selective-scan mixer $\mathcal{G}_{\mathrm{D}}$, which uses neither $S$ nor $\operatorname{LN}_{s}^{\mathrm{D}}$. Block $n$ applies a stack of $d_n$ DBSS modules independently to both streams:
\begin{equation}
\begin{split}
D_t^{i,n,0}&=X_t^{i,n-1},\\
D_t^{i,n,d_n}&=
\underbrace{\left(\mathcal{T}_{\mathrm{D}}^{n,d_n}\circ\cdots\circ
\mathcal{T}_{\mathrm{D}}^{n,1}\right)}_{d_n\ \mathrm{DBSS\ modules}}
\!\left(D_t^{i,n,0}\right),\quad t\in\{0,1\}.
\end{split}
\label{eq:dbssstack}
\end{equation}
After the final DBSS module, its spatial output is vectorised once to form the token sequence consumed by BiCA:
\begin{equation}
F_t^{i,n}=\operatorname{vec}(D_t^{i,n,d_n}),
\quad t\in\{0,1\}.
\label{eq:mv}
\end{equation}
The DBSS layer uses only the dual-branch mixer, without a hierarchical backbone, stage downsampling, or interleaved self-attention stages. Given a normalised feature map, an input projection creates two channel branches. The first branch passes through a depthwise convolution, SiLU activation, and selective state-space scan; the second follows a parallel non-scanned path with the same local convolutional preprocessing. Their outputs are concatenated and projected back to width $C$, after which the residual connection and MLP in Eq.~\eqref{eq:template} complete the layer. Spatially, the feature map is partitioned into non-overlapping $w\times w$ windows, the $w^2$ tokens in each window are processed, and the windows are restored to their original locations. Thus every DBSS layer preserves feature width and spatial resolution and never imports evidence from the other temporal image.

We use several DBSS layers when $d_n>1$ because one layer provides only one round of content-dependent propagation within each window. Stacking them deepens the processing within each window before BiCA compares the two dates.

BiCA reuses the complete module structure of Eq.~\eqref{eq:template}. It retains the pre-normalised mixer residual, channel-wise layer scale, pre-normalised MLP residual, and feature width of DBSS, and replaces only $\mathcal{G}_{\mathrm{D}}$ with a multi-head cross-attention mixer $\mathcal{G}_{\mathrm{B}}$. The source argument $S$ then turns the single-stream structure into a cross-stream module. Given $(F_0^{i,n},F_1^{i,n})$, the same BiCA module is invoked under two reciprocal query--source assignments. In direction $t\leftarrow1-t$, stream $t$ supplies the queries and stream $1-t$ supplies the keys and values. Both directions use the same pre-update pair and share $W^Q$, $W^K$, $W^V$, and $W^P$. Their updates return to the corresponding query streams, and the module thereby produces two directed outputs without fusion. For head $a$, let $Q_0^{(a)}$ and $[K_1^{(a)},V_1^{(a)}]$ be the $a$-th $C/h$-channel chunks of $\operatorname{LN}_Q(F_0^{i,n})W^Q$ and $\operatorname{LN}_{KV}(F_1^{i,n})[W^K,W^V]$. Direction $0\leftarrow1$ is
\begin{equation}
\begin{aligned}
H_{0\leftarrow1}^{(a)}&=\operatorname{Softmax}\!\left(
\frac{Q_0^{(a)}K_1^{(a)\top}}{\sqrt{C/h}}\right)V_1^{(a)},\\
A_{0\leftarrow1}&=\operatorname{Concat}_{a=1}^{h}
H_{0\leftarrow1}^{(a)}W^P,
\end{aligned}
\label{eq:bica1}
\end{equation}
where $a\in\{1,\ldots,h\}$. The two pre-projection layer normalisations have separate parameters. $A_{1\leftarrow0}$ is obtained by swapping the streams. In general, $A_{t\leftarrow1-t}$ is the evidence that stream $1-t$ contributes to stream $t$. The same cross-attention module is invoked in both directions. Weight sharing removes branch capacity as an explanation for directional differences. Unequal outputs can then be attributed to the ordered image content and its role as query or source under the shared mapping. $W^P$ is the output projection. The token sequence is then mapped back to the spatial grid by $\operatorname{vec}^{-1}$, the inverse of the vectorisation in Eq.~\eqref{eq:mv}.

Substituting $A_{t\leftarrow1-t}$ for the mixer output in Eq.~\eqref{eq:template} gives the BiCA update
\begin{equation}
\begin{split}
Z_t^{i,n}&=F_t^{i,n}+\gamma_1\odot A_{t\leftarrow 1-t},\\
X_t^{i,n}&=\operatorname{vec}^{-1}\!\Big(Z_t^{i,n}
+\gamma_2\odot\operatorname{MLP}\big(\operatorname{LN}(Z_t^{i,n})\big)\Big),
\end{split}
\label{eq:bica2}
\end{equation}
where $\gamma_1,\gamma_2\in\mathbb{R}^{C}$ are layer-scale parameters initialised to $10^{-5}$, $\odot$ is an elementwise product that rescales every channel independently, and the MLP expands to $4C$ with GELU. Each temporal stream serves once as the query, producing the updated pair returned by the block. Stochastic depth and dropout are disabled throughout the encoder. Every block holds its own DBSS and cross-attention parameters, leaving the $N$ blocks untied. The ordering gives DBSS responsibility for intra-image contextual refinement and BiCA responsibility for direct cross-temporal comparison.

Figure~\ref{fig:mvbica} illustrates the pair-preserving DS-BiCA encoder. Figure~\ref{fig:mvbica}(a) cascades pair-preserving blocks, bilinearly resamples all $2N$ temporal outputs to a common resolution, concatenates them, and projects, gates and normalises the result into the scale representation $Z^i$. The DS-BiCA blocks are shared across the three scales, while the aggregation head is scale-specific. Figure~\ref{fig:mvbica}(b) shows block $n$, which applies a shared DBSS stack of depth $d_n$ to both streams and BiCA in the two temporal directions, each producing $A_{t\leftarrow1-t}$ from Eq.~\eqref{eq:bica1}. The residual and MLP paths of Eq.~\eqref{eq:bica2} are internal to that module, and the inverse vectorisation follows it.

Figure~\ref{fig:mvbica}(c) labels the DBSS mixer parameters. Here $d_{\mathrm{state}}$, $d_{\mathrm{conv}}$, the expansion factor, and $r$ denote the state size, convolution width, channel expansion, and time-step rank. $W_{\mathrm{in}}$ forms the scanned and symmetric branches, $W_{\mathrm{scan}}$ produces $(\Delta,\boldsymbol B,\boldsymbol C)$, $W_{\Delta}$ expands the time-step rank, and $W_{\mathrm{out}}$ restores width $C$. In the selective scan, $\boldsymbol A$ is the state-transition matrix, $\boldsymbol B$ and $\boldsymbol C$ are input-dependent state maps, $\boldsymbol D$ is the skip coefficient, and $\Delta$ is the positive discretisation step~\cite{gu2024mamba}. The bold symbols are local to the mixer and differ from the batch size $B$ and common width $C$.

Figure~\ref{fig:mvbica}(d) labels the BiCA parameters. $W^Q$, $W^K$, $W^V$, and $W^P$ are the query, key, value, and output projections. $\operatorname{LN}_Q$ and $\operatorname{LN}_{KV}$ are separate pre-projection normalisations, while optional per-head query and key normalisations are identities. Across Figures~\ref{fig:mvbica}(c) and (d), SiLU and GELU are the displayed nonlinearities, and softplus makes $\Delta$ positive.

\subsection{Block and Scale Aggregation}\label{sec:aggregation}
All three scales share the $N$ encoder blocks and one set of encoder parameters for $i=2,3,4$, leaving only the subsequent aggregation scale-specific. One set of block parameters must therefore serve resolutions of $32^2$, $16^2$ and $8^2$, which constrains the encoder to scale-agnostic behaviour and keeps its parameter count independent of the number of scales. We then aggregate all block outputs. At scale $i$, every block output from both temporal streams is bilinearly resampled to a common resolution, which upsamples the mid-level and deep scales and leaves the shallow scale unchanged. The results are then concatenated along the channel dimension,
\begin{equation}
G^i=\Big[\operatorname*{Concat}_{n=1}^{N}\operatorname{up}(X_0^{i,n}),\ \operatorname*{Concat}_{n=1}^{N}\operatorname{up}(X_1^{i,n})\Big].
\label{eq:aggregate}
\end{equation}
A scale-specific linear projection restores width $C$. A gated linear unit (GLU), residual connection, and layer normalisation produce the representation that the decoder reads
\begin{equation}
\begin{aligned}
\widetilde G^i&=\operatorname{Linear}_i(G^i),\\
Z^i&=\operatorname{LN}_i\left(\widetilde G^i+\operatorname{GLU}\!\left(\operatorname{Linear}^{\mathrm{GLU}}_i(\widetilde G^i)\right)\right).
\end{aligned}
\label{eq:screp}
\end{equation}
$G^i$ has width $2NC$, $\operatorname{Linear}_i$ maps $2NC\to C$, and $\operatorname{Linear}^{\mathrm{GLU}}_i$ maps $C\to2C$ before the GLU halves the width back to $C$, giving $Z^i$ width $C$ at the common resolution. Both linear projections and the layer normalisation $\operatorname{LN}_i$ are scale-specific and not shared. Concatenation preserves temporal direction and block depth until the learned projection.

\subsection{Three-branch Gated Caption Decoder}
The decoder follows the design of ICT-Net~\cite{cai2023interactive}. At decoding position $j$, masked self-attention over the embedded prefix, with sinusoidal positional encoding and a post-norm residual, first produces a textual state $E^*$. A single cross-attention module is then queried once per scale representation. The three branches share one set of attention parameters and differ only in the evidence they read.
\begin{equation}
S_i=\operatorname{MHA}(E^*,Z^i,Z^i),\qquad i\in\{2,3,4\},
\label{eq:decoderattn}
\end{equation}
where $\operatorname{MHA}$ is multi-head attention with query, key and value in that order.
Each retrieved feature is controlled by a gate conditioned jointly on text and vision, where $\sigma$ is the logistic sigmoid. A single gate matrix $W_g\in\mathbb{R}^{C\times 2C}$ is shared by the three branches.
\begin{equation}
\widehat S_i=\sigma\!\left(W_g[E^*;S_i]\right)\odot S_i.
\label{eq:decodergate}
\end{equation}
The three gated outputs and the textual state are concatenated, projected from $4C$ to width $C$, and processed by a residual feed-forward network (FFN) of inner width $4C$ with ReLU.
\begin{equation}
\begin{aligned}
H&=\operatorname{LN}\!\left(U+\operatorname{FFN}(U)\right),\\
U&=\operatorname{LN}\!\left(W_o[\widehat S_2;\widehat S_3;\widehat S_4;E^*]\right).
\end{aligned}
\label{eq:decoderfusion}
\end{equation}
A final linear layer and softmax predict the next token.

During training, padded caption indices $Y\in\mathbb{N}^{B\times T}$ are word-embedded with sinusoidal positions and processed under an additive causal mask $M_{\mathrm c}\in\{0,-\infty\}^{T\times T}$. Here $T$ is the padded caption length. The vocabulary projection produces logits $O\in\mathbb{R}^{B\times T\times|\mathcal V|}$. With $m_{b,t}=1$ for a predicted non-padding token and $0$ otherwise, the captioning objective is
\begin{equation}
\begin{aligned}
\ell_{b,t}&=-\log p\!\left(Y_{b,t}\mid Y_{b,<t},I_0,I_1\right),\\
\mathcal{L}_{\mathrm{cap}}&=
\frac{\sum_{b=1}^{B}\sum_{t=1}^{T}m_{b,t}\ell_{b,t}}
{\sum_{b=1}^{B}\sum_{t=1}^{T}m_{b,t}}.
\end{aligned}
\label{eq:captionloss}
\end{equation}
All trainable components are jointly optimised using caption cross-entropy over non-padding target tokens, without label smoothing or auxiliary supervision.

\section{Experiments}\label{experiments}
\subsection{Datasets}

\textit{LEVIR-CC.} The dataset spans 5--14 years and contains 10,077 image pairs, split into 6,815 training, 1,333 validation, and 1,929 test pairs. Of these, 5,038 are changed and 5,039 unchanged. Each pair is $256\times256$ pixels with five captions. We retain words occurring at least five times~\cite{liu2022remote,cai2023interactive}, giving $|\mathcal V|=468$.

\textit{Dubai-CC.} The dataset covers Dubai over 10 years and contains 500 pairs, split into 300 training, 50 validation, and 150 test pairs. Each $50\times50$ pair has five captions and is resized to $256\times256$~\cite{li2024inter}. We retain every observed token, giving $|\mathcal V|=333$.
\subsection{Evaluation Metrics}
For all compared methods, we report BLEU-1 through BLEU-4~\cite{papineni2002bleu}, METEOR~\cite{banerjee2005meteor}, ROUGE\_L~\cite{chin2004rouge}, CIDEr~\cite{vedantam2015cider}, and the composite $S_m^*$~\cite{zhang2021global}, the mean of BLEU-4, METEOR, ROUGE\_L and CIDEr. Higher is better.
\subsection{Implementation Details}\label{sec:impl}
We implement the network in PyTorch on NVIDIA H100 GPUs. The ResNet101 backbone~\cite{he2016resnet} starts from ImageNet-pretrained weights~\cite{deng2009imagenet}, while DBSS and BiCA are trained from scratch. All trainable parameters are optimised with Adam~\cite{kingma2014adam} with $\beta_1$ of 0.9, $\beta_2$ of 0.999, $\epsilon$ of $10^{-8}$ and zero weight decay, a batch size of 16, and at most 400 epochs. The learning rate starts at $1\times10^{-4}$ and is scaled by 0.7 whenever validation $S_m^*$ does not improve for three consecutive epochs. The best validation model is evaluated on the test split.

The common width $C$ is 512 and the decoder has one layer with eight heads, with dropout $0.5$ on the self-attention output, on each cross-attention output, inside the FFN, and before the vocabulary projection. The ResNet101 stem and Stage~1 are frozen and Stages~2--4 are fine-tuned.

DBSS uses $8\times8$ windows without spatial downsampling. Each module keeps width $C$, uses an MLP ratio of $4$, disables drop path, and its mixer has a state size $d_{\mathrm{state}}$ of 8, a convolution width $d_{\mathrm{conv}}$ of 3, an expansion factor of one, and a time-step rank $r$ of $\lceil C/16\rceil$, or 32. Each BiCA module uses eight heads of dimension 64 with QKV bias and zero attention and projection dropout. Both module types initialise their layer scales at $10^{-5}$. Block outputs are upsampled to $32\times32$ before scale-specific aggregation.

Captions hold at most 50 lexical tokens, and the start and end markers bring a full decode to 52 steps. Inference uses beam size 3~\cite{cai2023interactive,li2024inter}. Paired images receive identical transforms, and both datasets use colour jitter with brightness, contrast, and saturation of $\pm0.2$ and zero hue.

Dubai-CC has three protocols, referred to below by number. Protocol 1 uses no geometric transformation. Protocol 2 randomly selects one of eight transformations: the identity, rotations by $90^\circ$, $180^\circ$, and $270^\circ$, horizontal and vertical flips, and reflections across the main and anti-diagonals. Protocol 3, fourfold sampling, first adds horizontal-flip, vertical-flip, and combined horizontal-and-vertical-flip copies of every pair and then applies the same eight transformations. It expands the geometric views of the same image pairs without adding new scenes.

The two benchmarks are otherwise trained alike. LEVIR-CC uses no geometric augmentation, whereas Dubai-CC uses one of the three protocols above.
\begin{figure}[t]
\centering
\includegraphics[width=\columnwidth]{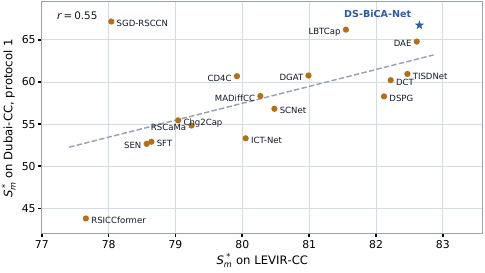}
\caption{$S_m^*$ of the sixteen methods that appear on both benchmarks, with our model at protocol 1 on Dubai-CC. The dashed line is the least-squares fit.}
\label{fig:crossdata}
\vspace{-2pt}
\end{figure}
\begin{table}[t]
  \caption{Performance comparison on the LEVIR-CC dataset.}
  \label{tab:levir}
  \centering
  \setlength{\tabcolsep}{1.5pt}\fontsize{7.0}{8.0}\selectfont
  \begin{tabular*}{\columnwidth}{@{\extracolsep{\fill}}lS[table-format=2.2]S[table-format=2.2]S[table-format=2.2]S[table-format=2.2]S[table-format=2.2]S[table-format=2.2]S[table-format=3.2]S[table-format=2.2]@{}}
    \toprule
    Method & {B-1} & {B-2} & {B-3} & {B-4} & {M} & {R-L} & {C} & {$S_m^*$} \\
    \midrule
\multicolumn{9}{@{}l}{\textit{Attention-based comparison}}\\
CD4C~\cite{li2025cd4c} & 86.74 & 78.71 & 71.48 & 65.41 & 40.56 & 75.72 & 138.00 & 79.92 \\
DetACC~\cite{li2024detection} & 87.26 & 79.16 & 71.74 & 65.56 & 40.65 & 75.97 & 139.99 & 80.54 \\
DSPG~\cite{li2026exploring} & 86.49 & 78.81 & 72.21 & 66.71 & 41.61 & 76.81 & 143.34 & 82.12 \\
SGD-RSCCN~\cite{sun2025scene} & 84.17 & 75.16 & 68.05 & 62.48 & 39.18 & 74.24 & 136.27 & 78.04 \\
RSICCformer~\cite{liu2022remote} & 84.72 & 76.27 & 68.87 & 62.77 & 39.61 & 74.12 & 134.12 & 77.66 \\
Chg2Cap~\cite{chang2023changes} & 86.14 & 78.08 & 70.66 & 64.39 & 40.03 & 75.12 & 136.61 & 79.04 \\
ICT-Net~\cite{cai2023interactive} & 86.06 & 78.12 & 71.45 & 66.12 & 40.51 & 75.21 & 138.36 & 80.05 \\
MAF-Net~\cite{chen2024multi} & 84.20 & 76.21 & 69.65 & 64.38 & 39.58 & 74.19 & 134.61 & 78.19 \\
MfrNet~\cite{xu2024mfrnet} & 85.86 & 78.29 & 72.13 & 67.21 & 40.22 & 75.19 & 139.56 & 80.55 \\
TISDNet~\cite{li2024inter} & 86.79 & 79.35 & 73.10 & \underline{67.93} & 41.60 & 76.64 & 143.71 & 82.47 \\
DAE~\cite{xian2025dynamic} & 87.40 & \underline{79.99} & \textbf{73.71} & \textbf{68.52} & 41.40 & 76.53 & 144.01 & \underline{82.61} \\
DGAT~\cite{qin2025dgat} & 87.09 & {--} & {--} & 65.30 & 40.98 & 76.41 & 141.27 & 80.99 \\
STAND~\cite{gong2026stand} & \textbf{88.24} & \textbf{80.33} & \underline{73.15} & 67.11 & 41.55 & 77.25 & 143.39 & 82.33 \\
SEN~\cite{zhou2024single} & 85.10 & 77.05 & 70.01 & 64.09 & 39.59 & 74.57 & 136.02 & 78.57 \\
SFT~\cite{sun2024lightweight} & 84.56 & 75.87 & 68.64 & 62.87 & 39.93 & 74.69 & 137.05 & 78.64 \\
SCNet~\cite{sun2026scnet} & 86.14 & 78.19 & 71.44 & 65.82 & 40.51 & 75.37 & 140.23 & 80.48 \\
CASC~\cite{wang2025change} & 87.03 & 79.56 & 73.13 & 67.86 & 41.81 & 76.42 & 140.97 & 81.77 \\
DCT~\cite{wang2025lightweight} & 86.79 & 79.39 & 72.10 & \underline{67.93} & 41.60 & 76.64 & 142.70 & 82.22 \\
LBTCap~\cite{zhang2026lbtcap} & 86.61 & 78.77 & 71.75 & 65.94 & 40.96 & 76.74 & 142.57 & 81.55 \\
\addlinespace[1pt]
\addlinespace[1pt]
\addlinespace[1pt]
\multicolumn{9}{@{}l}{\textit{Diffusion decoder}}\\
MADiffCC~\cite{yang2024remote} & 86.28 & 77.50 & 71.09 & 66.93 & 40.16 & 75.37 & 138.61 & 80.27 \\
\addlinespace[1pt]
\multicolumn{9}{@{}l}{\textit{Vision-language decoder}}\\
RSCC-Diff~\cite{yang2026rscc} & 86.81 & 78.49 & 71.16 & 65.12 & \underline{41.90} & 76.87 & \textbf{144.53} & 82.10 \\
Semantic-CC~\cite{zhu2024semantic} & \underline{88.07} & 79.68 & 71.47 & 64.51 & 40.58 & \underline{77.76} & 138.51 & 80.34 \\
KCFI~\cite{yang2025enhancing} & 86.34 & 77.31 & 70.89 & 65.30 & 39.42 & 75.47 & 138.25 & 79.61 \\
ChangeVG~\cite{xue2026towards} & {--} & {--} & {--} & 65.08 & \textbf{42.06} & 76.95 & 142.83 & 81.73 \\
CCExpert~\cite{wang2024ccexpert} & 86.65 & 78.47 & 71.31 & 65.49 & 41.82 & 76.55 & 143.32 & 81.80 \\
LLaVA-RS~\cite{yildirim2025llava} & {--} & {--} & {--} & 65.60 & 41.80 & \textbf{78.70} & 141.50 & 81.90 \\
\addlinespace[1pt]
\multicolumn{9}{@{}l}{\textit{State-space comparison}}\\
RSCaMa~\cite{liu2024rscama} & 85.79 & 77.99 & 71.04 & 65.24 & 39.91 & 75.24 & 136.56 & 79.24 \\
VMamba-CC~\cite{wang2025remote} & 84.37 & 76.35 & 68.96 & 62.67 & 39.01 & 74.26 & 132.83 & 77.19 \\
IHM-SNet~\cite{ma2025ihm} & 86.21 & 78.23 & 71.26 & 65.59 & 40.15 & 75.42 & 137.92 & 79.77 \\
MTI-CC~\cite{ma2025cross} & 86.34 & 78.64 & 71.81 & 66.14 & 40.26 & 75.54 & 138.82 & 80.19 \\
MTH-Net~\cite{ma2025mth} & 87.09 & 78.91 & 71.92 & 66.13 & 40.57 & 75.79 & 140.19 & 80.67 \\
RMNet~\cite{cao2026rmnet} & 86.09 & 78.43 & 71.68 & 66.11 & 40.94 & 75.57 & 139.43 & 80.51 \\
\midrule
    DS-BiCA-Net (\textbf{Ours}) & 86.90 & 79.40 & 73.00 & 67.82 & 41.68 & 77.01 & \underline{144.09} & \textbf{82.65} \\
\bottomrule
  \end{tabular*}
\vspace{-7pt}
\end{table}
\subsection{Baselines}\label{sec:baselines}
We compare DS-BiCA-Net with published RSICC methods in four groups. The attention-based group comprises RSICCformer~\cite{liu2022remote}, Chg2Cap~\cite{chang2023changes}, ICT-Net~\cite{cai2023interactive}, MAF-Net~\cite{chen2024multi}, MfrNet~\cite{xu2024mfrnet}, TISDNet~\cite{li2024inter}, SEN~\cite{zhou2024single}, SFT~\cite{sun2024lightweight}, DetACC~\cite{li2024detection}, CD4C~\cite{li2025cd4c}, SGD-RSCCN~\cite{sun2025scene}, DAE~\cite{xian2025dynamic}, DGAT~\cite{qin2025dgat}, CASC~\cite{wang2025change}, DCT~\cite{wang2025lightweight}, TACC~\cite{hang2025text}, DSPG~\cite{li2026exploring}, STAND~\cite{gong2026stand}, SCNet~\cite{sun2026scnet} and LBTCap~\cite{zhang2026lbtcap}. The state-space group comprises RSCaMa~\cite{liu2024rscama}, VMamba-CC~\cite{wang2025remote}, IHM-SNet~\cite{ma2025ihm}, MTI-CC~\cite{ma2025cross}, MTH-Net~\cite{ma2025mth} and RMNet~\cite{cao2026rmnet}. The diffusion group is represented by MADiffCC~\cite{yang2024remote}, and the vision-language group comprises Semantic-CC~\cite{zhu2024semantic}, Change-Agent~\cite{liu2024change}, CCExpert~\cite{wang2024ccexpert}, KCFI~\cite{yang2025enhancing}, LLaVA-RS~\cite{yildirim2025llava}, ChangeVG~\cite{xue2026towards} and RSCC-Diff~\cite{yang2026rscc}. Tables~\ref{tab:levir} and~\ref{tab:dubai} list each method only on the datasets for which it reports results.

\subsection{Quantitative Results}
\subsubsection{Results on the LEVIR-CC dataset}
Table~\ref{tab:levir} reports our method against published methods, grouped by paradigm. Values are percentages, bold marks the best and underline the second best, and B-$n$ denotes BLEU-$n$, M METEOR, R-L ROUGE\_L, and C CIDEr.

DS-BiCA-Net obtains the highest $S_m^*$ (82.65) and the second-highest CIDEr (144.09) among the listed methods. Baseline scores are quoted from their articles and we hold no trained model or per-image prediction for any of them. Among the listed Mamba-based approaches, DS-BiCA-Net is highest on every metric except BLEU-1, where MTH-Net is higher by 0.19 points. The MTH-Net result uses auxiliary change-mask supervision through joint change detection and caption training.

\subsubsection{Results on the Dubai-CC dataset}
Table~\ref{tab:dubai} reports our method under the three protocols of Section~\ref{sec:impl} against published methods. Protocol 1 applies no geometric transformation and is therefore the closest comparison we can offer. Under protocol 1, our model reaches $S_m^*=66.72$, the second-highest among the published results compared, and a CIDEr of 120.12. Protocol 2 raises $S_m^*$ to 70.14 and protocol 3 to 72.13. The latter two use geometric augmentation and are listed below the last rule, outside the ranking. Methods without published Dubai-CC values are omitted, and the metric abbreviations follow Table~\ref{tab:levir}. Among the available state-space baselines, RSCaMa reports Dubai-CC results, and DS-BiCA-Net under protocol 1 is higher on all seven metrics.

Figure~\ref{fig:crossdata} plots the sixteen methods that report on both benchmarks against each other. Their scores show a moderate cross-benchmark association, with Pearson $r=0.55$. SGD-RSCCN is second from last on LEVIR-CC and the highest baseline on Dubai-CC.

\begin{table}[t]
  \caption{Performance comparison on the Dubai-CC dataset. Bold and underline rank the published methods together with protocol 1. Protocols 2 and 3 use geometric augmentation and are listed below the last rule for reference only.}
  \label{tab:dubai}
  \centering
  \setlength{\tabcolsep}{1.5pt}\fontsize{6.3}{7.2}\selectfont
  \begin{tabular*}{\columnwidth}{@{\extracolsep{\fill}}lS[table-format=2.2]S[table-format=2.2]S[table-format=2.2]S[table-format=2.2]S[table-format=2.2]S[table-format=2.2]S[table-format=3.2]S[table-format=2.2]@{}}
    \toprule
    Method & {B-1} & {B-2} & {B-3} & {B-4} & {M} & {R-L} & {C} & {$S_m^*$} \\
    \midrule
\addlinespace[1pt]
\multicolumn{9}{@{}l}{\textit{Attention-based comparison}}\\
TACC~\cite{hang2025text} & 72.17 & 58.65 & 48.24 & 38.59 & 28.17 & 58.93 & 95.45 & 55.29 \\
CD4C~\cite{li2025cd4c} & 74.13 & 63.11 & 53.13 & 43.55 & 30.27 & 60.25 & 108.66 & 60.68 \\
DSPG~\cite{li2026exploring} & 72.34 & 60.52 & 50.30 & 40.65 & 29.40 & 62.66 & 100.42 & 58.28 \\
SGD-RSCCN~\cite{sun2025scene} & \textbf{77.15} & \textbf{66.80} & \textbf{58.07} & \textbf{50.27} & \textbf{34.07} & 65.88 & 118.42 & \textbf{67.16} \\
RSICCformer~\cite{liu2022remote} & 67.92 & 53.61 & 41.37 & 31.28 & 25.41 & 51.96 & 66.54 & 43.80 \\
SFT~\cite{sun2024lightweight} & 67.30 & 55.97 & 47.00 & 37.30 & 26.32 & 56.38 & 91.59 & 52.90 \\
Chg2Cap~\cite{chang2023changes} & 72.04 & 60.18 & 50.84 & 41.70 & 28.92 & 58.66 & 92.49 & 55.44 \\
ICT-Net~\cite{cai2023interactive} & 69.38 & 57.03 & 46.50 & 36.17 & 26.78 & 57.31 & 92.97 & 53.31 \\
TISDNet~\cite{li2024inter} & 71.81 & 60.60 & 50.88 & 41.37 & 29.67 & 63.24 & 109.47 & 60.94 \\
DAE~\cite{xian2025dynamic} & \underline{76.08} & 64.77 & 54.95 & 44.76 & 32.38 & \textbf{68.06} & 113.94 & 64.78 \\
DGAT~\cite{qin2025dgat} & 75.91 & 63.49 & 53.18 & 43.63 & 30.63 & 62.94 & 105.85 & 60.76 \\
SEN~\cite{zhou2024single} & 70.95 & 57.28 & 45.81 & 36.25 & 26.62 & 55.95 & 91.77 & 52.65 \\
SCNet~\cite{sun2026scnet} & 73.48 & 60.98 & 50.51 & 40.80 & 29.62 & 59.06 & 97.74 & 56.81 \\
DCT~\cite{wang2025lightweight} & 73.48 & 61.46 & 52.25 & 45.41 & 30.67 & 63.24 & 101.47 & 60.20 \\
LBTCap~\cite{zhang2026lbtcap} & 75.20 & 64.00 & 54.57 & 45.90 & \underline{33.25} & 66.86 & \underline{118.81} & 66.20 \\
\addlinespace[1pt]
\addlinespace[1pt]
\addlinespace[1pt]
\multicolumn{9}{@{}l}{\textit{Diffusion decoder}}\\
MADiffCC~\cite{yang2024remote} & 73.18 & 61.36 & 52.25 & 45.41 & 30.85 & 60.56 & 96.47 & 58.32 \\
\addlinespace[1pt]
\multicolumn{9}{@{}l}{\textit{Vision-language decoder}}\\
Change-Agent~\cite{liu2024change} & 73.24 & 61.17 & 51.03 & 40.19 & 28.73 & 60.38 & 101.77 & 57.77 \\
\multicolumn{9}{@{}l}{\textit{State-space comparison}}\\
RSCaMa~\cite{liu2024rscama} & 71.33 & 58.61 & 46.99 & 36.42 & 28.72 & 59.86 & 94.31 & 54.83 \\
\midrule
    DS-BiCA-Net (no geo.) & 75.43 & \underline{65.32} & \underline{56.22} & \underline{47.78} & 31.73 & \underline{67.27} & \textbf{120.12} & \underline{66.72} \\
    \midrule
    DS-BiCA-Net (8 transforms) & 74.70 & 64.86 & 56.95 & 49.09 & 33.06 & 69.80 & 128.62 & 70.14 \\
    DS-BiCA-Net (4$\times$ sampling) & 77.09 & 67.25 & 59.11 & 51.12 & 35.25 & 70.32 & 131.81 & 72.13 \\
\bottomrule
  \end{tabular*}
\vspace{-7pt}
\end{table}
\subsubsection{Accuracy and capacity trade-off}
\begin{figure}[t]
\centering
\includegraphics[width=\columnwidth]{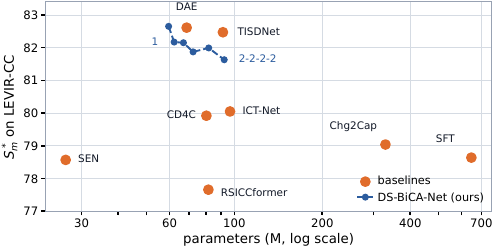}
\caption{Accuracy and capacity trade-off on LEVIR-CC.}
\label{fig:tradeoff}
\vspace{-2pt}
\end{figure}

Figure~\ref{fig:tradeoff} plots $S_m^*$ against parameter count on LEVIR-CC. The single-block configuration uses fewer parameters than every baseline except SEN, which it exceeds by $4.08$ $S_m^*$ points. It also reaches a higher $S_m^*$ than DAE and TISDNet, the two strongest baselines, at $59.58$M against $68.68$M and $91.40$M.

\subsubsection{Comparison of parameters and computational cost}\label{sec:cost}

DS-BiCA-Net can vary in the number of blocks and in the number of DBSS modules per block, and we name each configuration by its per-block DBSS counts joined by hyphens. For example, ``1'' is one block with one module and ``2-2-2-2'' is four blocks of two. We locally count parameters for every baseline and configuration in Table~\ref{tab:latency}. For our configurations, FLOPs include every forward-path component and use one $256\times256$ pair with a 52-token teacher-forced decoder pass. The fixed $42.50$M backbone dominates capacity, while encoder growth produces totals of 59.58 to 92.33M.

Table~\ref{tab:latency} reports end-to-end autoregressive latency on one NVIDIA H100 80\,GB HBM3 GPU and two 16-core Intel Xeon Gold 6326 processors with 64\,GB of host memory. Each is timed sequentially at batch size one on a bi-temporal pair, using fixed-length greedy decoding for the listed number of steps. Timing covers the complete forward path and excludes data loading, image decoding, host-to-device transfer, and metric computation. Latency grows from a mean of $52.70$\,ms for configuration 1 to $73.95$\,ms for 2-2-2-2.

At the same input resolution and decode length, configuration 1 matches the latency of ICT-Net, $52.70$ against $53.04$\,ms, with $38\%$ fewer parameters and $2.60$ more points of $S_m^*$. It is smaller than every baseline except SEN, at $59.58$\,M against $96.64$\,M for ICT-Net, $129.18$\,M for RMNet, $328.89$\,M for Chg2Cap and $898.93$\,M for RSCaMa. SEN alone is both smaller and faster, at $26.48$\,M and $26.57$\,ms, and configuration 1 exceeds it by $4.08$ points.

\begin{table}[t]
\caption{Parameter-count and end-to-end latency comparison. ``Pair'' is the input resolution and ``FPS'' is $1000/\text{mean}$.}
\label{tab:latency}
\centering
\setlength{\tabcolsep}{1.5pt}\fontsize{9.4}{10.8}\selectfont
\begin{tabular*}{\columnwidth}{@{\extracolsep{\fill}}lcS[table-format=2.0]S[table-format=3.2]S[table-format=2.2(2.2),separate-uncertainty=true]S[table-format=2.1]@{}}
\toprule
Method & Pair & {Steps} & {\textit{Para.} (M)} & {Mean $\pm$ std. (ms)} & {FPS} \\
\midrule
    SEN~\cite{zhou2024single} & $256^2$ & 52 & 26.48 & 26.57(1.24) & 37.6 \\
    Chg2Cap~\cite{chang2023changes} & $256^2$ & 41 & 328.89 & 38.63(1.26) & 25.9 \\
    ICT-Net~\cite{cai2023interactive} & $256^2$ & 52 & 96.64 & 53.04(0.70) & 18.9 \\
    RSCaMa~\cite{liu2024rscama} & $224^2$ & 42 & 898.93 & 31.05(1.06) & 32.2 \\
    RMNet~\cite{cao2026rmnet} & $224^2$ & 52 & 129.18 & 59.90(17.01) & 16.7 \\
    \midrule
    \multicolumn{6}{@{}l}{DS-BiCA-Net (\textbf{Ours})} \\
    \quad 1 & $256^2$ & 52 & 59.58 & 52.70(2.45) & 19.0 \\
    \quad 2 & $256^2$ & 52 & 62.23 & 54.57(2.43) & 18.3 \\
    \quad 1-1 & $256^2$ & 52 & 66.96 & 57.17(2.98) & 17.5 \\
    \quad 2-2 & $256^2$ & 52 & 72.26 & 61.40(3.24) & 16.3 \\
    \quad 1-1-1-1 & $256^2$ & 52 & 81.72 & 65.58(2.30) & 15.3 \\
    \quad 2-2-2-2 & $256^2$ & 52 & 92.33 & 73.95(3.07) & 13.5 \\
\bottomrule
\end{tabular*}
\vspace{-7pt}
\end{table}
\subsection{Qualitative Results}
\subsubsection{LEVIR-CC caption comparison}
\begin{figure}[t]
\centering
\fontsize{6.5}{7.5}\selectfont\setlength{\baselineskip}{6.3pt}
\settowidth{\gtw}{\textbf{GT.} }
\setlength{\tabcolsep}{0pt}
\renewcommand{\arraystretch}{1.00}
\begin{tabular}{@{}p{0.485\linewidth}@{\hspace{0.03\linewidth}}p{0.485\linewidth}@{}}
\toprule
\begin{minipage}[t]{\linewidth}
{\raggedright\includegraphics[width=0.36\linewidth]{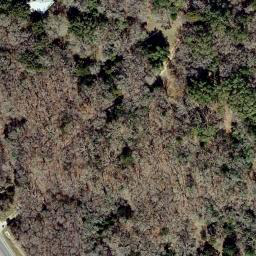}\hspace{4pt}\includegraphics[width=0.36\linewidth]{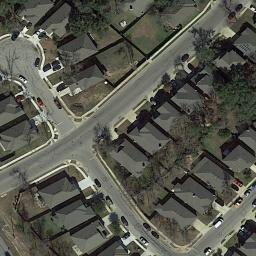}\par}
\vspace{4pt}
\raggedright\qhang\textbf{GT.} a residential area with many houses and roads appears\qcont the woods become a complex of buildings with crossroads in between\qcont several concrete roads and many houses have been constructed to replace the original trees\qcont the forest is replaced by some roads with many houses on both sides\qcont massive houses along the roads replace the woods \par\vspace{2.5pt}\textbf{ICT-Net.} many \textcolor[HTML]{1E7A3C}{houses} are \textcolor[HTML]{C0392B}{built} \textcolor[HTML]{1E7A3C}{along} the \textcolor[HTML]{1E7A3C}{road} in the \textcolor[HTML]{C0392B}{open} \textcolor[HTML]{C0392B}{space} \qhang\textbf{Chg2Cap.} many \textcolor[HTML]{1E7A3C}{houses} are \textcolor[HTML]{C0392B}{built} \textcolor[HTML]{1E7A3C}{along} the \textcolor[HTML]{1E7A3C}{roads} \qhang\textbf{SEN.} many \textcolor[HTML]{1E7A3C}{houses} are \textcolor[HTML]{C0392B}{built} on both \textcolor[HTML]{1E7A3C}{sides} of the \textcolor[HTML]{1E7A3C}{road} \qhang\textbf{RSCaMa.} many \textcolor[HTML]{1E7A3C}{houses} are \textcolor[HTML]{C0392B}{built} \textcolor[HTML]{1E7A3C}{along} the \textcolor[HTML]{1E7A3C}{road} \qhang\textbf{Ours.} a \textcolor[HTML]{1E7A3C}{residential} \textcolor[HTML]{1E7A3C}{area} with many \textcolor[HTML]{1E7A3C}{houses} and \textcolor[HTML]{1E7A3C}{roads} \textcolor[HTML]{1E7A3C}{appears}
\end{minipage} &
\begin{minipage}[t]{\linewidth}
{\raggedright\includegraphics[width=0.36\linewidth]{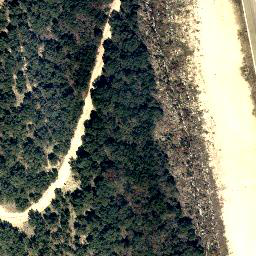}\hspace{4pt}\includegraphics[width=0.36\linewidth]{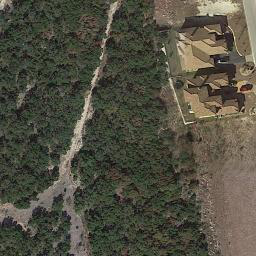}\par}
\vspace{4pt}
\raggedright\qhang\textbf{GT.} two villas have been built on the bareland near the forest\qcont two yellow houses are situated in front of the forest on the ground\qcont two houses appear in the desert at the edge of the forest\qcont two villas appear on the right of the forest\qcont two buildings appear on the bareland in the right \par\vspace{2.5pt}\textbf{ICT-Net.} some \textcolor[HTML]{1E7A3C}{houses} are \textcolor[HTML]{1E7A3C}{built} at the \textcolor[HTML]{C0392B}{top} of the \textcolor[HTML]{C0392B}{scene} \qhang\textbf{Chg2Cap.} a \textcolor[HTML]{C0392B}{row} of \textcolor[HTML]{1E7A3C}{villas} is \textcolor[HTML]{1E7A3C}{built} on the \textcolor[HTML]{1E7A3C}{right} \textcolor[HTML]{C0392B}{side} of the \textcolor[HTML]{C0392B}{scene} \qhang\textbf{SEN.} a \textcolor[HTML]{C0392B}{row} of \textcolor[HTML]{1E7A3C}{houses} is \textcolor[HTML]{1E7A3C}{built} \textcolor[HTML]{C0392B}{along} the \textcolor[HTML]{C0392B}{road} \qhang\textbf{RSCaMa.} \textcolor[HTML]{C0392B}{three} \textcolor[HTML]{1E7A3C}{houses} are \textcolor[HTML]{1E7A3C}{built} in the \textcolor[HTML]{C0392B}{woods} \qhang\textbf{Ours.} \textcolor[HTML]{1E7A3C}{two} \textcolor[HTML]{1E7A3C}{houses} are \textcolor[HTML]{1E7A3C}{built} in the \textcolor[HTML]{1E7A3C}{forest}
\end{minipage} \\
\midrule
\begin{minipage}[t]{\linewidth}
{\raggedright\includegraphics[width=0.36\linewidth]{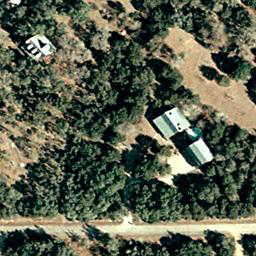}\hspace{4pt}\includegraphics[width=0.36\linewidth]{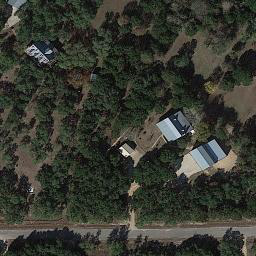}\par}
\vspace{4pt}
\raggedright\qhang\textbf{GT.} a small house has appeared in the trees\qcont a house appears in the center of the forest\qcont a small house is built in the center of the scene\qcont a separate building shows up in the forest\qcont a building appears in the middle of the scene \par\vspace{2.5pt}\textbf{ICT-Net.} the \textcolor[HTML]{1E7A3C}{scene} is the \textcolor[HTML]{C0392B}{same} as \textcolor[HTML]{C0392B}{before} \qhang\textbf{Chg2Cap.} the \textcolor[HTML]{1E7A3C}{scene} is the \textcolor[HTML]{C0392B}{same} as \textcolor[HTML]{C0392B}{before} \qhang\textbf{SEN.} a \textcolor[HTML]{1E7A3C}{house} is \textcolor[HTML]{1E7A3C}{built} in the \textcolor[HTML]{C0392B}{woods} \qhang\textbf{RSCaMa.} the \textcolor[HTML]{1E7A3C}{scene} is the \textcolor[HTML]{C0392B}{same} as \textcolor[HTML]{C0392B}{before} \qhang\textbf{Ours.} a \textcolor[HTML]{1E7A3C}{small} \textcolor[HTML]{1E7A3C}{building} \textcolor[HTML]{1E7A3C}{appears} in the \textcolor[HTML]{1E7A3C}{forest}
\end{minipage} &
\begin{minipage}[t]{\linewidth}
{\raggedright\includegraphics[width=0.36\linewidth]{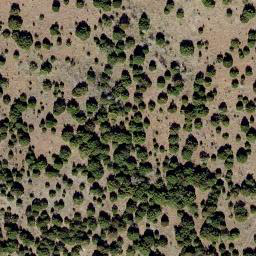}\hspace{4pt}\includegraphics[width=0.36\linewidth]{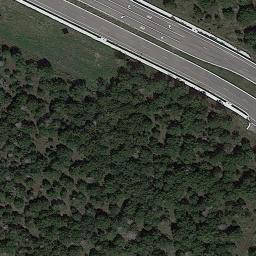}\par}
\vspace{4pt}
\raggedright\qhang\textbf{GT.} the woods become denser and two roads appear on the upper right\qcont two concrete roads are built and the vegetation on this land has increased\qcont two straight roads are built at the top parallel\qcont a road appears at the top right corner of the scene\qcont two roads appear among the plants \par\vspace{2.5pt}\textbf{ICT-Net.} a \textcolor[HTML]{1E7A3C}{road} is \textcolor[HTML]{1E7A3C}{built} \textcolor[HTML]{C0392B}{across} the \textcolor[HTML]{C0392B}{forest} \qhang\textbf{Chg2Cap.} a \textcolor[HTML]{1E7A3C}{road} \textcolor[HTML]{1E7A3C}{appears} in the \textcolor[HTML]{C0392B}{center} of the \textcolor[HTML]{C0392B}{forest} \qhang\textbf{SEN.} a \textcolor[HTML]{1E7A3C}{road} is \textcolor[HTML]{1E7A3C}{built} in the \textcolor[HTML]{1E7A3C}{woods} \qhang\textbf{RSCaMa.} a \textcolor[HTML]{1E7A3C}{road} \textcolor[HTML]{1E7A3C}{appears} at the \textcolor[HTML]{1E7A3C}{top} \textcolor[HTML]{1E7A3C}{right} \textcolor[HTML]{1E7A3C}{corner} of the \textcolor[HTML]{1E7A3C}{scene} \qhang\textbf{Ours.} a \textcolor[HTML]{1E7A3C}{road} \textcolor[HTML]{1E7A3C}{appears} in the \textcolor[HTML]{C0392B}{desert}
\end{minipage} \\
\bottomrule
\end{tabular}
\caption{Qualitative comparison on the LEVIR-CC test set.}
\label{fig:caption_results}
\vspace{-2pt}
\end{figure}

Figure~\ref{fig:caption_results} compares configuration 1 on LEVIR-CC with four baselines~\cite{cai2023interactive,chang2023changes,zhou2024single,liu2024rscama}. Each case contains the pair, five references, and predictions. The first three show successes and the fourth a failure. A content word is green when it occurs in a reference and red otherwise, and function words are black.
The successes cover scene-level replacement, a small isolated change, and an explicit count. The last example is retained because a higher composite score did not prevent a localisation error on that pair.

\subsubsection{Dubai-CC training-protocol comparison}
\begin{figure}[t]
\centering
\fontsize{6.5}{7.5}\selectfont\setlength{\baselineskip}{6.3pt}
\settowidth{\gtw}{\textbf{GT.} }
\noindent\begin{minipage}[t]{0.485\columnwidth}
\textbf{(a) Missed change becomes a road description}\par
\includegraphics[width=0.36\linewidth]{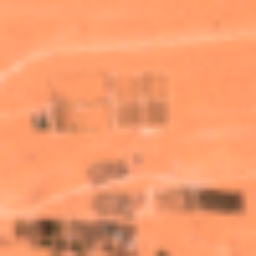}\hspace{4pt}\includegraphics[width=0.36\linewidth]{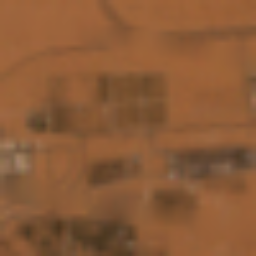}\par
\raggedright\qhang\textbf{GT.} More buildings and new streets are present in the desert.\par
\qcont A street has been built.\par
\qcont A road has been created in the desert.\par
\qcont Some new houses and a road appear in the desert.\par
\qcont A road was edified above the houses.\par
\labline{no geo. (1).}{\textcolor[HTML]{C0392B}{Everything} \textcolor[HTML]{C0392B}{remains} the \textcolor[HTML]{C0392B}{same}.}
\labline{8 transforms (1).}{The \textcolor[HTML]{C0392B}{area} was \textcolor[HTML]{C0392B}{urbanized}.}
\labline{4$\times$ sampling (2-2-2-2).}{A \textcolor[HTML]{1E7A3C}{road} was \textcolor[HTML]{1E7A3C}{built} in the \textcolor[HTML]{1E7A3C}{desert}.}
\end{minipage}\hfill
\begin{minipage}[t]{0.485\columnwidth}
\textbf{(b) Progressively more specific urban change}\par
\includegraphics[width=0.36\linewidth]{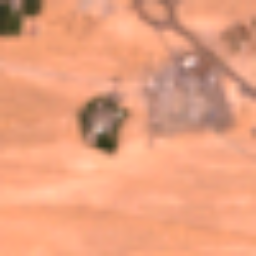}\hspace{4pt}\includegraphics[width=0.36\linewidth]{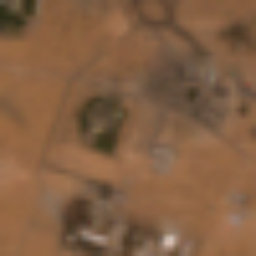}\par
\raggedright\qhang\textbf{GT.} Some roads and constructions were created near the existing ones.\par
\qcont The area was urbanized.\par
\qcont Some houses were added near old ones.\par
\qcont Some buildings have been built in the desert.\par
\qcont Many new buildings and new roads appear in the desert.\par
\labline{no geo. (1).}{The \textcolor[HTML]{1E7A3C}{area} \textcolor[HTML]{1E7A3C}{appears} the \textcolor[HTML]{C0392B}{same}.}
\labline{8 transforms (1).}{The \textcolor[HTML]{1E7A3C}{area} was \textcolor[HTML]{1E7A3C}{urbanized}.}
\labline{4$\times$ sampling (2-2-2-2).}{Some \textcolor[HTML]{1E7A3C}{buildings} were \textcolor[HTML]{1E7A3C}{built} \textcolor[HTML]{1E7A3C}{near} the \textcolor[HTML]{1E7A3C}{existing} \textcolor[HTML]{1E7A3C}{ones}.}
\end{minipage}

\vspace{3pt}
\noindent\begin{minipage}[t]{0.485\columnwidth}
\textbf{(c) Richer artificial-island description}\par
\includegraphics[width=0.36\linewidth]{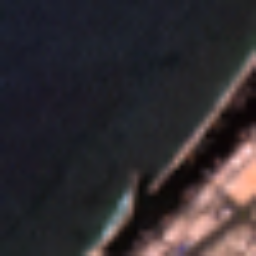}\hspace{4pt}\includegraphics[width=0.36\linewidth]{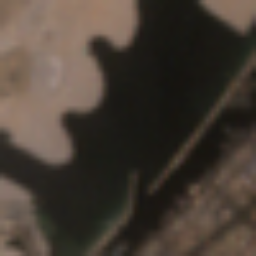}\par
\raggedright\qhang\textbf{GT.} One big artificial island made of dredged sand and rock appears near the coast.\par
\qcont A big artificial island has been created in the sea.\par
\qcont An artificial island made of dredged sand and rock appears near the shore.\par
\qcont One island was built on the left of the shore.\par
\qcont An artificial island appeared in the sea.\par
\labline{no geo. (1).}{An \textcolor[HTML]{1E7A3C}{artificial} \textcolor[HTML]{1E7A3C}{island} was \textcolor[HTML]{1E7A3C}{created} in the \textcolor[HTML]{1E7A3C}{sea}.}
\labline{8 transforms (1).}{An \textcolor[HTML]{1E7A3C}{artificial} \textcolor[HTML]{1E7A3C}{island} was \textcolor[HTML]{1E7A3C}{built} in the \textcolor[HTML]{1E7A3C}{sea}.}
\labline{4$\times$ sampling (2-2-2-2).}{An \textcolor[HTML]{1E7A3C}{artificial} \textcolor[HTML]{1E7A3C}{island} \textcolor[HTML]{1E7A3C}{made} of \textcolor[HTML]{1E7A3C}{dredged} \textcolor[HTML]{1E7A3C}{sand} and \textcolor[HTML]{1E7A3C}{rock} was \textcolor[HTML]{1E7A3C}{built} in the \textcolor[HTML]{1E7A3C}{sea}.}
\end{minipage}\hfill
\begin{minipage}[t]{0.485\columnwidth}
{\textbf{(d) Challenging urban-change case}}\par
\includegraphics[width=0.36\linewidth]{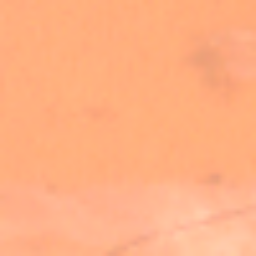}\hspace{4pt}\includegraphics[width=0.36\linewidth]{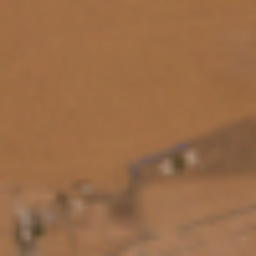}\par
\raggedright\qhang\textbf{GT.} Some constructions appear in the desert.\par
\qcont Some new roads and some constructions have been realized in the desert.\par
\qcont Some buildings and streets were realized in the desert.\par
\qcont New buildings and roads were created in the desert.\par
\qcont Some buildings have been edified in the desert.\par
\labline{no geo. (1).}{Some \textcolor[HTML]{1E7A3C}{buildings} were \textcolor[HTML]{C0392B}{built} in the \textcolor[HTML]{1E7A3C}{desert}.}
\labline{8 transforms (1).}{Some \textcolor[HTML]{1E7A3C}{buildings} were \textcolor[HTML]{C0392B}{built} in the \textcolor[HTML]{1E7A3C}{desert}.}
\labline{4$\times$ sampling (2-2-2-2).}{A \textcolor[HTML]{C0392B}{small} \textcolor[HTML]{C0392B}{park} was \textcolor[HTML]{C0392B}{built} in the \textcolor[HTML]{1E7A3C}{desert}.}
\end{minipage}
\vspace{4pt}
\caption{Dubai-CC qualitative comparison across the three training protocols.}
\label{fig:dubai_protocols}
\vspace{-2pt}
\end{figure}

Figure~\ref{fig:dubai_protocols} compares the three Dubai-CC protocols on four fixed pairs with shared references, coloured as in Figure~\ref{fig:caption_results}. Figures~\ref{fig:dubai_protocols}(a) to (c) show the description growing more specific from protocol 1 to protocol 3, matching the ranking of the three protocols in Table~\ref{tab:dubai}, and Figure~\ref{fig:dubai_protocols}(d) a case that none of the three protocols gets right.

\subsection{Word-level Attention Analysis}
\begin{figure}[t]
\centering
{\footnotesize\textbf{(a) correctly grounded}}\\[1pt]
\includegraphics[width=\columnwidth]{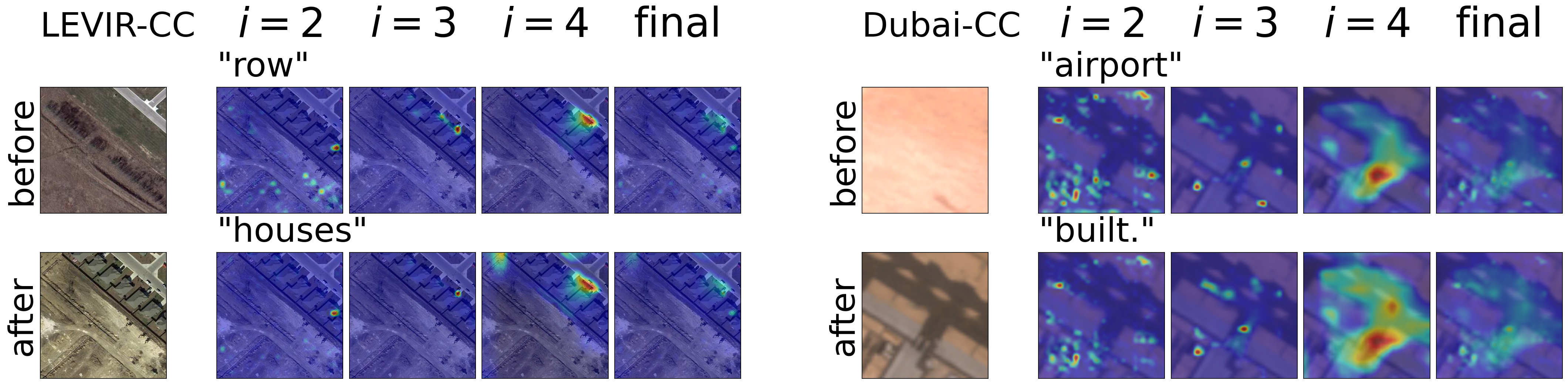}\\[0pt]
\includegraphics[width=\columnwidth]{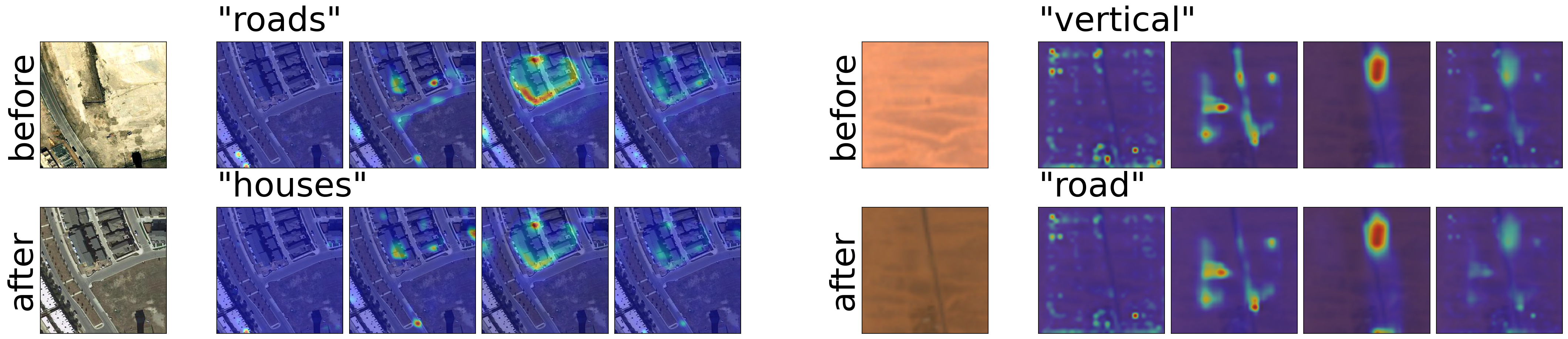}\\[0pt]
\includegraphics[width=\columnwidth]{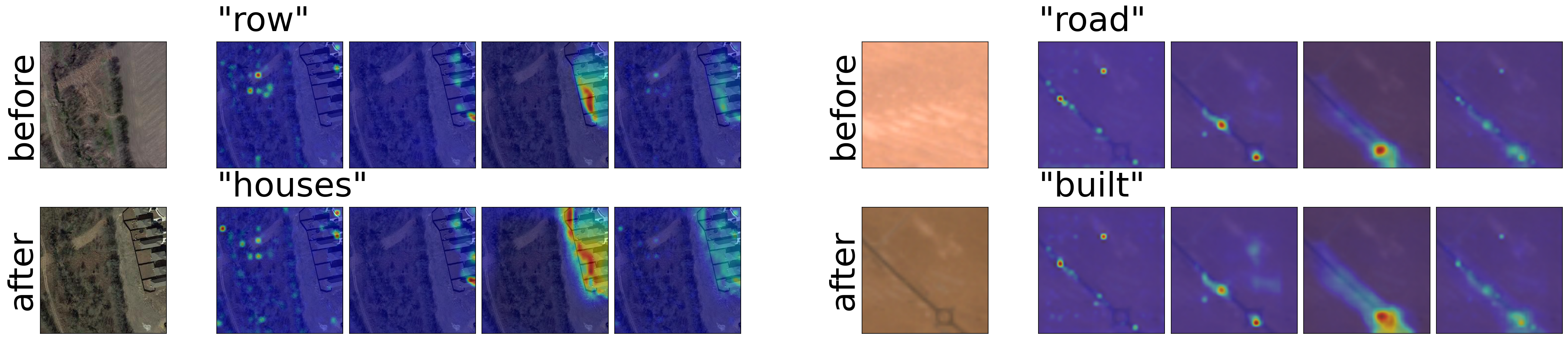}\\[3pt]
{\footnotesize\textbf{(b) challenging cases}}\\[1pt]
\includegraphics[width=\columnwidth]{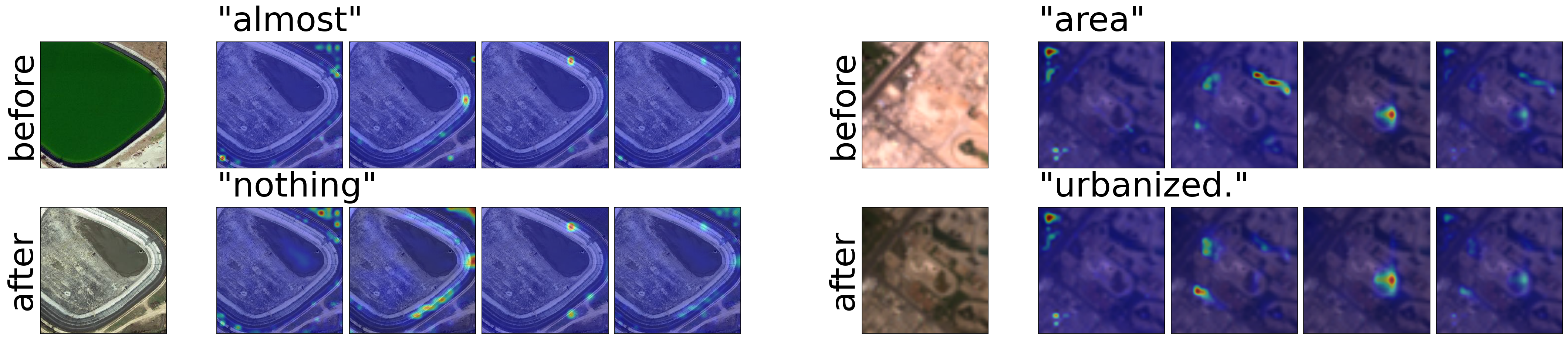}
\caption{Word-level decoder cross-attention decomposed by scale.}
\label{fig:attention}
\vspace{-2pt}
\end{figure}

Figure~\ref{fig:attention} decomposes head-averaged decoder cross-attention by word and scale ($i=2,3,4$), with the final column the mean of the three min-max-normalised scale maps. Figure~\ref{fig:attention}(a) shows successful cases and Figure~\ref{fig:attention}(b) challenging cases from both datasets. Comparing the three scales, deep-scale maps are more spatially coherent and shallow-scale maps respond more diffusely to texture and edges. The difference between scales is visible because Eq.~\eqref{eq:decoderattn} computes a separate attention softmax for each scale.

\subsection{Failure Analysis}
Table~\ref{tab:failure_analysis} audits all 1,929 LEVIR-CC and 150 Dubai-CC test predictions against five references after merging inflectional variants. The audited models are the single-block configuration on LEVIR-CC and the 2-2-2-2 configuration trained under protocol 3 on Dubai-CC. An object, position, or count is reference-supported if mentioned by any reference. A wrong object replaces a supported object with another category. An unsupported-object hallucination is any unsupported substituted or additional object. A wrong count conflicts with a supported count for the same object, and a wrong position conflicts with a supported spatial direction. A missed change omits a supported event or reports no change. A pseudo-change predicts an event when all references report none. A small-change miss requires ``small'', ``tiny'', or ``little'' in a reference and does not measure area. Categories may overlap. Entries are sample counts with percentages, and lower is better. Small-change misses lead on LEVIR-CC. Missed changes lead on Dubai-CC.

\begin{table}[t]
  \caption{Caption-level failure analysis on LEVIR-CC and Dubai-CC.}
  \label{tab:failure_analysis}
  \centering
  \setlength{\tabcolsep}{1.5pt}\fontsize{9.2}{10.6}\selectfont
  \begin{tabular*}{\columnwidth}{@{\extracolsep{\fill}}lcc@{}}
    \toprule
    Error type & LEVIR-CC & Dubai-CC \\
    \midrule
    Small-change miss & 89 (4.61\%) & 5 (3.33\%) \\
    Wrong object & 7 (0.36\%) & 1 (0.67\%) \\
    Wrong object position & 0 (0.00\%) & 0 (0.00\%) \\
    Wrong count & 40 (2.07\%) & 0 (0.00\%) \\
    Pseudo-change & 28 (1.45\%) & 3 (2.00\%) \\
    Unsupported-object hallucination & 79 (4.10\%) & 4 (2.67\%) \\
    Missed change & 80 (4.15\%) & 15 (10.00\%) \\
    \bottomrule
  \end{tabular*}
\vspace{-7pt}
\end{table}

\subsection{Ablation Studies}
Three questions about the encoder can be settled by removal, namely whether DBSS and BiCA each work on their own, whether the effect of one depends on the presence of the other, and whether the order in which they are applied matters. The rows of Table~\ref{tab:ablation} name these variants. Identity keeps neither operator, DBSS-only keeps DBSS alone, BiCA-only keeps the cross-attention alone, and BiCA$\rightarrow$DBSS applies the two in the reverse order. A fourth control, written UniCA, keeps the shared BiCA projections but retains a single temporal direction, giving each single-direction variant the same parameter count as the bidirectional module. UniCA $1\!\leftarrow\!0$ retains only $A_{1\leftarrow0}$ of Eq.~\eqref{eq:bica1}, in which $I_1$ queries $I_0$, and UniCA $0\!\leftarrow\!1$ retains only $A_{0\leftarrow1}$. A factorial design answers the first two together, since removing one module at a time leaves an interaction invisible. Two further ablations vary the capacity axes of the encoder, the number of DBSS modules per block and the number of blocks.

In Table~\ref{tab:ablation} the recurring ``1'' row is at once the full factorial cell, the DBSS-first order, and the single-block configuration. Costs follow Section~\ref{sec:cost}. Recurring architectures therefore share FLOPs, only vocabulary size separates the two parameter counts, and retaining one direction halves the arithmetic of UniCA.
\subsubsection{Factorial ablation of DBSS and BiCA}
The $2\times2$ factorial rows of Table~\ref{tab:ablation}(a) and (b) retain ResNet101 and vary only DBSS and BiCA under matched colour-jitter-only training. Identity contains no interaction encoder and therefore isolates what the shared backbone, aggregation, and decoder achieve without encoder-level temporal interaction. Removing a component also removes its parameters, and we do not compensate for the loss, evaluating each variant at whatever size removal leaves it.

Whether DBSS helps depends on what follows it. On its own it lowers the score on both benchmarks, and it raises the score only once BiCA follows. Its effect thus reverses sign with the presence of the cross-temporal operator. The reversal holds across two benchmarks that differ by a factor of twenty in training pairs and cover different sensors and scenes, and averaging the two cases would report an operator that does nothing. At a matched parameter count neither single direction recovers the bidirectional result on either benchmark. Adding one direction to DBSS improves on DBSS alone in both orders on Dubai-CC, but in only one of the two orders on LEVIR-CC. What DBSS needs after it is therefore the pair of directed comparisons rather than any cross-temporal operator. Identity reaches 82.04 on LEVIR-CC, showing that removing the interaction encoder does not remove downstream processing of the paired backbone features, since the aggregation head still forms the scale representations the decoder requires.

The result is the design rationale of Section~\ref{method} put to test. Intra-image context has no consumer that can use it for comparison until BiCA follows, at which point each query reaches the cross-temporal match already carrying a spatially coherent view of its own image.

The evidence has limits. The full model stands above BiCA-only by $0.44$ points on LEVIR-CC and $1.72$ on Dubai-CC, the same direction on both, and Table~\ref{tab:paired} places neither gap outside the sampling noise of these test-set sizes. The two benchmarks thus fix the direction of the conditional gain but leave its size open.
Neither parameter-matched UniCA direction reaches BiCA on either dataset, and the directions differ from each other.

\begin{figure}[t]
\centering
\includegraphics[width=\columnwidth]{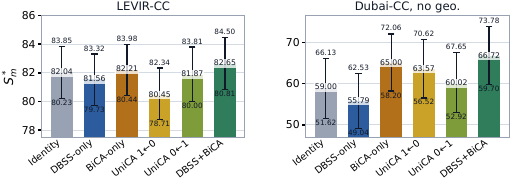}
\caption{Factorial and direction controls from Table~\ref{tab:ablation}(a) and (b). Bars show complete-test scores with marginal 95\% bootstrap intervals. Paired tests appear in Table~\ref{tab:paired}.}
\label{fig:ablation}
\vspace{-2pt}
\end{figure}

Figure~\ref{fig:ablation} reports the sampling uncertainty of the factorial and direction ablations. For every model in Table~\ref{tab:ablation}(a) and (b), we resample the test images with replacement 1,000 times under a fixed seed~\cite{efron1993bootstrap}. BLEU-4 is recomputed from its corpus sufficient statistics in each replicate instead of being averaged over images~\cite{koehn2004statistical}. METEOR, ROUGE\_L and CIDEr are recomputed on the resampled prediction and reference pairs using their standard corpus scorers, and $S_m^*$ is recalculated from the four. The intervals on Dubai-CC are about four times wider than those on LEVIR-CC, which is consistent with its substantially smaller test split of 150 images, compared with 1,929 for LEVIR-CC.

Table~\ref{tab:paired} therefore uses identical indices for 1,000 paired replicates, two-sided centred bootstrap $p$ values~\cite{koehn2004statistical}, and the Holm correction over the ten contrasts~\cite{holm1979simple}. Every row contrasts the full model with one ablated variant, and a positive $\Delta S_m^*$ favours the full model. Two of the ten contrasts survive the correction at the $0.05$ level, the full model over UniCA $1\!\leftarrow\!0$ on LEVIR-CC and over DBSS-only on Dubai-CC, and both are marked in bold. Neither benchmark separates the factorial interaction from zero. The ten contrasts form one family across both benchmarks, and correcting each benchmark as its own family of five leaves the same two significant.

\begin{table}[t]
\caption{Ablation studies. Metric abbreviations follow Table~\ref{tab:levir}.}
\label{tab:ablation}
\centering
\scriptsize
\setlength{\tabcolsep}{1.5pt}
\resizebox{\columnwidth}{!}{%
\begin{tabular}{@{}lS[table-format=2.2]S[table-format=2.2]S[table-format=2.2]S[table-format=2.2]S[table-format=2.2]S[table-format=2.2]S[table-format=3.2]S[table-format=2.2]S[table-format=2.2]S[table-format=3.2]@{}}
\toprule
Setting and model & {B-1} & {B-2} & {B-3} & {B-4} & {M} & {R-L} & {C} & {$S_m^*$} & {\textit{Para.} (M)} & {FLOPs (G)} \\
\midrule
\multicolumn{11}{@{}l}{(a) LEVIR-CC}\\
\quad Identity & 87.05 & 79.02 & 72.03 & 66.30 & 41.59 & 76.79 & 143.46 & 82.04 & 53.78 & 26.70 \\
\quad DBSS-only & 86.92 & 79.20 & 72.56 & 67.23 & 41.24 & 76.18 & 141.59 & 81.56 & 56.43 & 33.82 \\
\quad BiCA-only & \textbf{87.53} & \textbf{79.45} & 72.13 & 65.94 & 41.27 & \textbf{77.20} & \textbf{144.45} & 82.21 & 56.93 & 37.45 \\
\quad UniCA $1\!\leftarrow\!0$ & 85.32 & 77.63 & 71.16 & 65.94 & 40.50 & 75.45 & 139.91 & 80.45 & 59.58 & 39.20 \\
\quad UniCA $0\!\leftarrow\!1$ & 86.48 & 79.04 & 72.57 & 67.23 & 41.19 & 76.29 & 142.76 & 81.87 & 59.58 & 39.20 \\
\quad BiCA$\rightarrow$DBSS & 86.62 & 78.82 & 72.12 & 66.76 & 41.38 & 76.76 & 142.30 & 81.80 & 59.58 & 44.57 \\
\quad 1 & 86.90 & 79.40 & \textbf{73.00} & \textbf{67.82} & \textbf{41.68} & 77.01 & 144.09 & \textbf{82.65} & 59.58 & 44.57 \\
\quad 2 & 86.97 & 79.25 & 72.59 & 67.19 & 41.62 & 76.76 & 143.09 & 82.17 & 62.23 & 51.69 \\
\quad 1-1 & 87.02 & 79.28 & 72.64 & 67.19 & 41.53 & 76.78 & 143.11 & 82.15 & 66.96 & 64.05 \\
\quad 2-2 & 85.96 & 78.30 & 71.78 & 66.46 & 41.36 & 76.51 & 143.16 & 81.87 & 72.26 & 78.28 \\
\quad 1-1-1-1 & 86.56 & 78.96 & 72.57 & 67.50 & 41.20 & 76.55 & 142.70 & 81.99 & 81.72 & 103.00 \\
\quad 2-2-2-2 & 86.90 & 78.97 & 72.11 & 66.51 & 41.43 & 76.51 & 142.05 & 81.63 & 92.33 & 131.46 \\
\midrule
\multicolumn{11}{@{}l}{(b) Dubai-CC, protocol 1}\\
\quad Identity & 66.17 & 57.54 & 49.90 & 42.00 & 28.70 & 59.71 & 105.57 & 59.00 & 53.64 & 26.70 \\
\quad DBSS-only & 68.67 & 57.47 & 48.02 & 38.98 & 27.70 & 59.54 & 96.94 & 55.79 & 56.29 & 33.82 \\
\quad BiCA-only & 74.12 & 62.89 & 53.83 & 45.48 & 31.28 & 65.41 & 117.84 & 65.00 & 56.79 & 37.45 \\
\quad UniCA $1\!\leftarrow\!0$ & 70.79 & 60.11 & 51.34 & 42.66 & 31.28 & 65.81 & 114.52 & 63.57 & 59.44 & 39.20 \\
\quad UniCA $0\!\leftarrow\!1$ & 69.38 & 59.52 & 51.70 & 44.20 & 29.27 & 61.00 & 105.62 & 60.02 & 59.44 & 39.20 \\
\quad BiCA$\rightarrow$DBSS & 71.24 & 60.31 & 51.71 & 43.56 & 31.25 & 65.86 & 116.57 & 64.31 & 59.44 & 44.57 \\
\quad 1 & 75.43 & \textbf{65.32} & \textbf{56.22} & \textbf{47.78} & 31.73 & 67.27 & \textbf{120.12} & \textbf{66.72} & 59.44 & 44.57 \\
\quad 2 & 72.01 & 60.80 & 51.43 & 42.15 & 30.60 & 63.23 & 109.56 & 61.39 & 62.10 & 51.69 \\
\quad 1-1 & 74.05 & 63.29 & 53.82 & 44.86 & \textbf{31.88} & 65.84 & 117.71 & 65.07 & 66.82 & 64.05 \\
\quad 2-2 & \textbf{76.09} & 65.01 & 54.65 & 44.43 & 31.86 & \textbf{68.19} & 117.90 & 65.60 & 72.13 & 78.28 \\
\quad 1-1-1-1 & 74.00 & 62.78 & 52.83 & 43.11 & 30.47 & 63.67 & 110.04 & 61.82 & 81.58 & 103.00 \\
\quad 2-2-2-2 & 74.63 & 62.51 & 51.74 & 41.04 & 31.39 & 66.62 & 112.18 & 62.81 & 92.19 & 131.46 \\
\midrule
\multicolumn{11}{@{}l}{(c) Dubai-CC, protocol 2}\\
\quad BiCA$\rightarrow$DBSS & 73.62 & 62.91 & 54.05 & 44.58 & 31.79 & 66.26 & 118.76 & 65.35 & 59.44 & 44.57 \\
\quad 1 & 74.70 & 64.86 & \textbf{56.95} & \textbf{49.09} & 33.06 & 69.80 & \textbf{128.62} & \textbf{70.14} & 59.44 & 44.57 \\
\quad 2 & 68.84 & 60.25 & 52.89 & 45.51 & 31.34 & 68.12 & 118.87 & 65.96 & 62.10 & 51.69 \\
\quad 1-1 & \textbf{74.96} & \textbf{65.18} & 56.31 & 47.29 & \textbf{33.51} & 68.36 & 126.17 & 68.83 & 66.82 & 64.05 \\
\quad 2-2 & 71.18 & 60.98 & 51.86 & 43.92 & 32.04 & 65.40 & 116.12 & 64.37 & 72.13 & 78.28 \\
\quad 1-1-1-1 & 74.57 & 63.95 & 54.39 & 44.28 & 32.38 & \textbf{70.06} & 119.20 & 66.48 & 81.58 & 103.00 \\
\quad 2-2-2-2 & 73.74 & 63.62 & 55.04 & 46.37 & 31.29 & 67.95 & 120.53 & 66.53 & 92.19 & 131.46 \\
\midrule
\multicolumn{11}{@{}l}{(d) Dubai-CC, protocol 3}\\
\quad 1 & 74.12 & 64.16 & 55.36 & 46.79 & 33.50 & 68.42 & 122.67 & 67.85 & 59.44 & 44.57 \\
\quad 2 & 75.47 & 66.57 & 59.38 & 52.20 & 34.56 & 68.87 & 130.65 & 71.57 & 62.10 & 51.69 \\
\quad 1-1 & 74.16 & 64.49 & 55.96 & 47.58 & 33.85 & 69.34 & 127.70 & 69.62 & 66.82 & 64.05 \\
    \quad 2-2 & 75.67 & 65.89 & 57.98 & 49.63 & 34.32 & \textbf{70.43} & 131.80 & 71.54 & 72.13 & 78.28 \\
\quad 1-1-1-1 & 75.35 & 66.82 & \textbf{60.11} & \textbf{53.61} & 34.95 & 67.81 & 129.59 & 71.49 & 81.58 & 103.00 \\
    \quad 2-2-2-2 & \textbf{77.09} & \textbf{67.25} & 59.11 & 51.12 & \textbf{35.25} & 70.32 & \textbf{131.81} & \textbf{72.13} & 92.19 & 131.46 \\
\bottomrule
\end{tabular}}
\vspace{-7pt}
\end{table}

\subsubsection{Module order of DBSS and BiCA}
Matched single-block comparisons in Table~\ref{tab:ablation}(a), (b), and (c) vary only module order, at the same parameter count and the same FLOPs. DBSS$\rightarrow$BiCA exceeds BiCA$\rightarrow$DBSS on every one of the seven metrics, in each of the three settings. On $S_m^*$ it leads by $0.85$ points on LEVIR-CC, by $2.41$ on Dubai-CC protocol 1 and by $4.79$ on Dubai-CC protocol 2.

\subsubsection{Impact of DBSS module count}
At equal block count, one module per block leads throughout LEVIR-CC and mostly under Dubai-CC protocols 1 and 2. Two modules lead throughout protocol 3. Across Dubai-CC protocols, every configuration improves from protocol 1 to protocol 3, though not monotonically, since protocol 2 is a peak for the single-block model at $70.14$ and a dip for 2-2 at $64.37$.

\begin{figure}[t]
\centering
\setlength{\abovecaptionskip}{-2pt}
\includegraphics[width=\columnwidth]{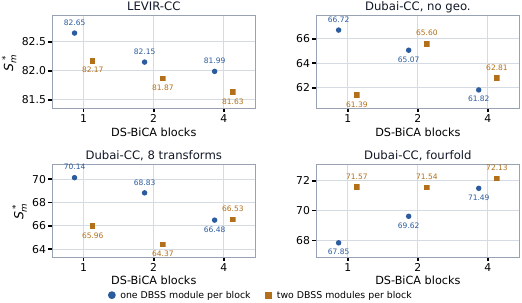}
\caption{Caption quality under the two encoder scaling axes.}
\label{fig:scaling}
\vspace{-2pt}
\end{figure}
\begin{table}[t]
\caption{Paired bootstrap comparisons against the single-block configuration, the best model on LEVIR-CC and on Dubai-CC protocol 1. $\Delta S_m^*$ is the difference of the complete-test-set scores in Table~\ref{tab:ablation}, while the interval and $p$ come from the bootstrap replicates. The last block reports the $2\times2$ factorial interaction, which is exploratory and outside the corrected family.}
\label{tab:paired}
\centering
  \setlength{\tabcolsep}{1.5pt}
\fontsize{8.1}{9.3}\selectfont
\begin{tabular*}{\columnwidth}{@{\extracolsep{\fill}}lS[table-format=2.2]cS[table-format=1.3]S[table-format=1.3]@{}}
\toprule
Contrast & {$\Delta S_m^*$} & 95\% CI & {$p$} & {Holm $p$} \\
\midrule
\multicolumn{5}{@{}l}{LEVIR-CC}\\
\quad Identity & 0.61 & $[-0.48, 1.57]$ & 0.230 & 0.919 \\
\quad DBSS-only & 1.09 & $[0.06, 2.15]$ & 0.038 & 0.252 \\
\quad BiCA-only & 0.44 & $[-0.54, 1.55]$ & 0.401 & 0.919 \\
\quad \textbf{UniCA $1\!\leftarrow\!0$} & \bfseries 2.20 & \textbf{[1.05, 3.42]} & \bfseries 0.001 & \bfseries 0.010 \\
\quad UniCA $0\!\leftarrow\!1$ & 0.78 & $[-0.24, 1.81]$ & 0.117 & 0.584 \\
\midrule
\multicolumn{5}{@{}l}{Dubai-CC}\\
\quad Identity & 7.72 & $[1.12, 14.12]$ & 0.019 & 0.152 \\
\quad \textbf{DBSS-only} & \bfseries 10.93 & \textbf{[5.31, 17.07]} & \bfseries 0.001 & \bfseries 0.010 \\
\quad BiCA-only & 1.72 & $[-2.55, 6.20]$ & 0.458 & 0.919 \\
\quad UniCA $1\!\leftarrow\!0$ & 3.15 & $[-2.45, 8.96]$ & 0.279 & 0.919 \\
\quad UniCA $0\!\leftarrow\!1$ & 6.70 & $[0.63, 12.94]$ & 0.036 & 0.252 \\
\midrule
\multicolumn{5}{@{}l}{Factorial interaction $\Delta_{\mathrm{int}}$}\\
\quad LEVIR-CC & 0.92 & $[-0.49, 2.47]$ & 0.204 & {--} \\
\quad Dubai-CC & 4.93 & $[-4.13, 13.60]$ & 0.273 & {--} \\
\bottomrule
\end{tabular*}
\vspace{-7pt}
\end{table}

\subsubsection{Influence of DS-BiCA block count}

Figure~\ref{fig:scaling} visualises the complete Table~\ref{tab:ablation} depth results. With one module per block, $S_m^*$ falls monotonically as the encoder grows from one to two to four blocks on LEVIR-CC and on Dubai-CC protocols 1 and 2, and rises monotonically under protocol 3. With two modules per block only the LEVIR-CC fall survives, and all three Dubai-CC protocols turn non-monotone, placing their best score at two blocks under protocol 1 and at four blocks under protocols 2 and 3. Protocol 3 is therefore the only setting that favours four blocks at both module counts, and the depth is not free, since four blocks carry $37\%$ more parameters and $2.3$ times the FLOPs of one block.
\subsection{Encoder Activity}
Table~\ref{tab:encact} verifies that training moves the encoder away from its $10^{-5}$ layer-scale initialisation. The learned scales grow by two to three orders of magnitude, while one block changes its input norm by $20.4\%$ on LEVIR-CC and $4.0\%$ per block on Dubai-CC. The $\gamma$ columns give mean absolute learned scales, ``$\times$ init'' gives their pooled ratio to the initial value, and $\Delta$ input is the relative feature change over 40 random test pairs, both streams, and all scales.

\begin{table}[t]
  \caption{Encoder activity in the two models.}
  \label{tab:encact}
  \centering
  \setlength{\tabcolsep}{1.5pt}\fontsize{7.5}{8.6}\selectfont
  \begin{tabular*}{\columnwidth}{@{\extracolsep{\fill}}lS[table-format=1.2e-1]S[table-format=1.2e-1]S[table-format=3.0]S[table-format=2.1]@{}}
    \toprule
    Model & {BiCA $\gamma$} & {DBSS $\gamma$} & {$\times$ init} & {$\Delta$ input (\%)} \\
    \midrule
    LEVIR-CC, $S_m^*$ $82.65$ & 9.96e-3 & 3.32e-3 & 664 & 20.4 \\
    Dubai-CC, $S_m^*$ $72.13$ & 3.43e-3 & 1.57e-3 & 219 & 4.0 \\
    \bottomrule
  \end{tabular*}
\vspace{-7pt}
\end{table}
\section{Conclusion}\label{conclusion}
We presented DS-BiCA-Net, a pair-preserving encoder for remote sensing change captioning that assigns intra-image and cross-temporal modeling to different operators. DBSS modules model the spatial context of each image, a BiCA module compares the two images in both temporal directions with shared parameters, and the two streams stay separate until the aggregation head. Experiments on LEVIR-CC and Dubai-CC show competitive accuracy, with DBSS helping only when BiCA follows it and with both directions outperforming either one alone.

\section*{Acknowledgment}
Dr.~Naveed Akhtar is a recipient of the Australian Research Council Discovery Early Career Researcher Award (project number DE230101058) funded by the Australian Government.

\FloatBarrier
\bibliographystyle{IEEEtran}
\bibliography{main}

\vspace{11pt}

\vspace{11pt}

\vfill

\end{document}